\documentclass[twoside,11pt]{article}

\usepackage[preprint]{jmlr2e}
\usepackage{amsmath,color,comment}
\usepackage{dsfont}
\usepackage{algorithm}
\usepackage{algpseudocode}
\usepackage{subfigure}
\usepackage{hyperref}
\hypersetup{
    colorlinks=true,
    citecolor=blue,    
    linkcolor=red,     
    urlcolor=cyan,     
}

\newtheorem{thm}{Theorem}
\newtheorem{lem}{Lemma}
\newtheorem{prop}{Proposition}

\newtheorem{defn}{Definition}

\providecommand{\mb}[1]{\boldsymbol{#1}}

\providecommand{\mbb}[1]{\mathbb{#1}}

\makeatletter
\def\blfootnote{\xdef\@thefnmark{}\@footnotetext}
\makeatother

\newcommand{\argmax}{\operatornamewithlimits{argmax}}

\title{Gaussian mixture models as a proxy for interacting language models}

\jmlrheading{23}{2025}{1-\pageref{LastPage}}{???; Revised ???}{???}{21-0000}{Wang and Sharifi et al.}

\ShortHeadings{GMMs as a proxy for interacting language models}{Wang and Sharifi et al.}
\firstpageno{1}

\begin{document}

\title{Gaussian mixture models as a proxy for interacting language models}

\author{\name Edward L. Wang$^*$ \email ewang43@jhu.edu \\
        \addr Department of Applied Mathematics and Statistics\\
        Johns Hopkins University\\
       Baltimore, MD 21218, USA
       \AND 
        \name Mohammad Sharifi Kiasari$^*$ \email sharifi@umd.edu \\
        \addr Department of Mathematics\\
       University of Maryland\\
       College Park, MD 20742, USA
       \AND
       \name Tianyu Wang \email twang147@jhu.edu \\
       \addr Department of Applied Mathematics and Statistics\\
        Johns Hopkins University\\
       Baltimore, MD 21218, USA
        \AND
       \name Hayden Helm \email hayden@helivan.io \\
       \addr Helivan Research\\
       San Francisco, CA, USA
       \AND
       \name Avanti Athreya \email dathrey1@jhu.edu \\
       \addr Department of Applied Mathematics and Statistics\\
        Johns Hopkins University\\
       Baltimore, MD 21218, USA
       \AND
       \name Carey Priebe \email cep@jhu.edu \\
       \addr Department of Applied Mathematics and Statistics\\
        Johns Hopkins University\\
       Baltimore, MD 21218, USA
       \AND
        \name Vince Lyzinski \email vlyzinsk@umd.edu\\
       \addr Department of Mathematics\\
       University of Maryland\\
       College Park, MD 20742, USA
}
\editor{------}

\maketitle
\blfootnote{$^*$ equal contributors }


\begin{abstract}
Large language models (LLMs) are powerful tools that, in a number of settings, overlap with the results of human pattern recognition and reasoning. Retrieval-augmented generation (RAG) further allows LLMs to produce tailored output depending on the contents of their RAG databases. However, LLMs depend on complex, computationally expensive algorithms. In this paper, we introduce interacting Gaussian mixture models (GMMs) as a proxy for interacting LLMs. We construct a model of interacting GMMs, complete with an analogue to RAG updating, under which GMMs can generate, exchange, and update data and parameters. We show that this interacting system of Gaussian mixture models, which can be implemented at minimal computational cost, mimics certain aspects of experimental simulations of interacting LLMs whose iterative responses depend on feedback from other LLMs. We build a Markov chain from this system of interacting GMMs; formalize and interpret the notion of polarization for such a chain; and prove lower bounds on the probability of polarization. This provides theoretical insight into the use of interacting Gaussian mixture models as a computationally efficient proxy for interacting large language models. 
\end{abstract}

\section{Introduction}
\label{sec:introduction}

Research and development of large language models (LLMs) has progressed considerably in the past several years, with the result that large language models can now replicate several aspects of human pattern recognition, reasoning, and behavior in linguistic, cognitive and social sciences \citep{aher2023using, cai2023does, qiu2023does}. Additionally, retrieval augmented generation (RAG) allows LLMs to store and retrieve relevant information from a database of information as additional context when completing a task, producing wider-ranging and more accurate responses \citep{lewis2020retrieval, gao2023retrieval}. Since the RAG database of information includes previous sources of signal that the language model has received and processed, the incorporation of RAGs holds promise for computational frameworks that effectively model social interactions between individuals with diverse prior experiences and perspectives. 

While in no way a substitute for human social networks, the construction of such computational frameworks can potentially yield useful insights 
\citep{helm2024tracking}.
In such settings, simulating human social networks via systems of interacting LLMs allows for the study of complex phenomena in settings where human data is unavailable or intractable \citep{abbasiantaeb2024let}.
In \cite{mcguinness2024investigating}, for example, the time-varying output of interacting LLM agents is used to model two well-documented features of human social networks: first, human mirroring \citep{iacoboni2009mirroring}, in which individuals copy the behaviors of others; and second, social alignment \citep[see, for example,][]{sa1,sa2}, which is the emergence of a common perspective among individuals. 
In \cite{mcguinness2024investigating}, the authors endow multiple LLMs, each representing a given agent, with RAG databases that are initialized with differing perspectives on a certain type of flower. Over a period of time, agents interact with each other, updating their RAG databases with responses from other agents. The collection of responses to the question of which flower ``is prettiest" is measured at each time step. 

Their approach yields conclusions matching those found in prior literature, along with new results on the effect of various parameters of social network evolution on social alignment \citep[see, for example,][]{levy2020social,xu2021polarization}. 
Of note are the formation of {\em stable silos}, in which an LLM's communication with others does not meaningfully alter said agent's own perspective, as well as {\em unstable silos}, which correspond to a dynamic equilibrium in which agents oscillate between a small number of different viewpoints. 
For the related notion of polarization dynamics in social networks, see, for example \cite{guerra2013measure, matakos2017measuring, bail2018exposure}.
In \cite{mcguinness2024investigating}, the authors further analyze the relationship of local and global communication patterns (i.e., restricting the communication neighborhood size for each LLM aagent) on the type and number of silos formed.  

Despite the immense utility of modern language models in social science research, their inner workings are largely a black box.  
Thorough understanding of these black box generative models is further obscured by differing language embedding functions \citep{nie2024text}, 
complexities of deep neural networks \citep{bianchini2014complexity}, 
varying architectures (including transformers \citep{vaswani2017attention}, diffusion \citep{li2022diffusion}, and state space \citep{wang2024state}), 
and massive quantities of data. This lack of standardization raises questions about the reliability of conclusions arising from complex systems of interacting LLMs. Furthermore, LLMs are computationally intensive, which translates to heavy energy, hardware, and time costs for training and text generation \citep{jiang2024preventing}. These concerns motivate us to consider whether we can develop alternative computational models that are simpler, interpretable, computationally inexpensive, and theoretically tractable but also capture enough of the complexity of interacting LLM systems to be useful.

In this paper, motivated by \cite{shumailov2023curse}, we propose a computational framework utilizing multiple Gaussian mixture models (GMMs) as interacting agents, each with an associated set of vectors to act as a RAG database.  For background on GMMs, see for example, \cite{reynolds2009gaussian,scrucca2016mclust}.
Gaussian mixture models are well studied, and we can consider a training mechanism in which beliefs or perspectives are updated via the Expectation-Maximization (EM) algorithm \citep{mclachlan2008algorithm}.  
We implement a system of interacting GMMs which generate, exchange, and update data and weights. Using this system, we run. similar experiments to those in \cite{mcguinness2024investigating}, comparing the results obtained from their framework of interacting LLMs and our interacting GMMs. We find through simulation that interacting GMMs can replicate the complex time varying dynamics, such as stable and unstable siloing, of interacting LLMs,  but with a more easily interpretable underlying algorithm and far less computational cost. We build a Markov chain from this system of interacting Gaussian mixture models and describe a straightforward mechanism for RAG updates. We formalize what it means for the chain to {\em polarize}, and we prove lower bounds, under certain conditions, on the probability of polarization. 

{\bf Outline of paper}. We organize the paper as follows. In Section \ref{sec:LLM}, we describe background on interaction large language models and summarize relevant prior results, and we define and describe specifics of how Gaussian mixture models interact, exchange data, and update their respective RAGs, as well as construct an associated Markov chain whose behavior we can analyze formally. In Section \ref{sec:silo} we discuss the notion of silos and define polarization, and we state our core theorems on the probability of polarization. In Section \ref{sec:results}, we present empirical results for our system of interacting GMMs and compare them to prior results in \cite{mcguinness2024investigating} for interacting LLMs, and we examine the impact of GMM parameters on the formation of silos. In Sections \ref{sec:discussion} and \ref{sec:conclusion}, we provide further analysis of our results, additional perspective, caveats, and directions for ongoing research. 
\section{Large language and Gaussian mixture models of interacting agents}
\label{sec:LLM}

{\bf Background: LLM simulations for interacting agents}. We briefly describe the LLM simulation methodology from \cite{mcguinness2024investigating} with which we compare our simulation. 
In experiments, \cite{mcguinness2024investigating} investigates the time-varying beliefs of agents by asking about preferences among LLM agents over a collection of flowers; in particular, preferences are captured by asking questions about each LLM agent's ``favorite'' or ``prettiest'' flower (flowers providing a fertile space for preference to be analyzed and captured). 

The LLM simulation consists of $n$ agents $\{A_i\}_{i=1}^m$ where the agents are LLaMA-2-7B-Chat models \citep{touvron2023llama}, each equipped with external RAG databases with a fixed number of sentences about flowers. 
Since these agents change over time, we denote the $i$th agent at time $t$ as $A_i^{(t)}$. 
Each LLM has its RAG database initialized with random sentences about flowers, generated by asking ChatGPT to ``describe the beauty of various flowers.''
At each time $t= 0, 1, \dots, T$, the agent is asked to describe the prettiest flower. For each agent $i = 1, \dots, n$, let $B_i^{(t)}$ represents the answer of agent $A_i^{(t)}$ to the prettiest flower query at time step $t$. 
First, the type of flower that the agent describes is extracted from the answer and converted to a natural number Flower ID. 
This is used to determine the silo that the agent is in at time $t$. Next, the response $B_i^{(t)}$ is embedded to a vector $X_i^{(t)}$ in $\mb{R}^{768}$ using nomic-embed-v1.5 \citep{nussbaum2024nomic}.
The distance between two LLM agents at time $t$ is defined as the $l_2$ distance between the embeddings of their responses; to wit, the distance between models $i$ and $j$ at time $t$ is then given by
\begin{equation}
   D_{ij}^{(t)}= d(A_i^{(t)}, A_j^{(t)}) = \big\|{X_i^{(t)}} - {X_j^{(t)}}\big\|_2.
\end{equation}
Note that measuring the difference between more complex query sets could be quantified by measuring the distance between models in Data Kernel Perspective Space \citep{helm2024statistical}.

For each $t = 1, \dots, T$, the LLM performs an interaction. For each agent $A_i^{(t-1)}$ (chosen in a random order), we pick another agent $A_j^{(t-1)}$ to interact with. 
With mirroring probability $p$, we let $j = i$, so that the agent interacts with itself. 
Otherwise, $j$ is randomly chosen from the $k$ nearest neighbor agents to $A_i^{(t-1)}$; nearest according to $D^{(t-1)}$. 
The distance between two LLM agents is defined as the $l_2$ distance between the embeddings of their responses.
Once $j$ is selected, agent $A_i^{(t-1)}$ updates its RAG with the new sentence $B_i^{(t)}$; this update could either be according to a first-in-last-out heuristic or by dropping the element of the RAG farthest from $B_i^{(t)}$. 
We then define $A_i^{(t)}$ to be the new agent with the updated RAG. 
Note that during the interaction process, the core LLMs are not updated, with the entirety of the update occurring in the RAGs.


\subsection{Gaussian mixture models and their RAGs}
\label{ssec:gmm-model}

For integer $m>0$, we define $[m]=\{1,2,\cdots,m\}$. Motivated by \cite{shumailov2023curse}, we model our interacting system of LLMs via a more tractable interacting system of GMMs.
In the GMM system at time $t$, we define $\{A_i^{(t)}\}_{i=1}^m$ to be the $m$ distinct agents; each agent $A_i^{(t)}$ is represented as a pair: a GMM model $\mathcal{F}_i^{(t)}$ and a ``RAG'' set $R_i^{(t)}$. 
We next describe these two elements and the means of interaction in the GMM network below.

Each GMM $\mathcal{F}_i^{(t)}$ is a mixture over $n$, $d$-dimensional Gaussian components, and is parameterized by
\begin{itemize}
    \item[i.] A set of mean vectors $\mb{M}_{i}^{(t)} = \{\vec\mu_{i1}^{(t)}, \dots, \vec\mu_{in}^{(t)}\}$ where each $\vec \mu_{ij}^{(t)}\in \mbb{R}^d$; note that in our model we have that $\mb{M}=\mb{M}_{j}^{(t)}=\mb{M}_{i}^{(s)}$ for all $i,j\in[m]$ and times $s,t>0$; i.e., the set of means is common for all models and does not change in time.
    If each mean represents a ``topic'' in the interacting LLM system, it is reasonable that the set of possible topics is common to all agents and does not vary in time.
    \item[ii.] A set of covariance matrices $\mb{V}_{i}^{(t)} = \{\Sigma_{i1}^{(t)}, \dots, \Sigma_{in}^{(t)}\}$ where each $\Sigma_{ij}^{(t)}\in \mbb{R}^{d\times d}$. 
    If these $\Sigma$s are capturing the variability or breadth of each ``topic'' in the interacting LLM system, then it would again be reasonable to have $\mb{V}=\mb{V}_{j}^{(t)}=\mb{V}_{i}^{(s)}$ for all $i,j\in[m]$ and times $s,t>0$.
    While we keep the $\mb{V}$s fixed for our theoretical developments, it is easy to implement an update mechanism for these covariances; see the Remark \ref{rem:sigma} in Section \ref{sec:interact} for detail.
    \item[iii.] A set of weight vectors $\mb{w}_i^{(t)} \in \triangle^{n-1}$ for each of the $m$ GMMs, respectively, at time $t$. 
    Each component of the weight vector describes the probability of a draw from a Gaussian centered at one of the $\vec\mu_j$'s and can be interpreted as the agent's degree of certainty in perspective $\vec\mu_j$. These weight vectors vary between agents and change over time, defining the agent's beliefs; see Section \ref{sec:interact} for detail.
\end{itemize}

\noindent Each GMM is also endowed with additional information meant to model the RAG in the LLM framework.
For agent $i$, the RAG set $R_i^{(t)}$ is a set of elements in $\mbb{R}^d$ with a fixed size $r$, representing the unique knowledge and memories of agent $A_i^{(t)}$ at time $t$.

\subsection{Modeling interactions between GMMs: initialization and updates}
\label{sec:interact}


Our interaction model, described in Algorithm \ref{alg:standard-simulation-procedure}, is comprised of two steps: initialization and interaction. We outline these below in Sections \ref{sec:init}-\ref{sec:interaction}. 
Note that in Algorithm \ref{alg:standard-simulation-procedure}, three key function primitives are called:
\begin{itemize}
    \item[i.] For integer $h>0$, \texttt{SampleFromGMM}($\mb{w}_i^{(t)}, \mb{M}_{i}^{(t)}, \mb{V}_{i}^{(t)}$, $h$) generates $h$ i.i.d. samples from the GMM with means $\mb{M}_{i}^{(t)}$, covariances $\mb{V}_{i}^{(t)}$, and weights $\mb{w}_{i}^{(t)}$.
    \item[ii.] \texttt{UpdateGMM}($R_i^{(t)},\mb{M}_{i}^{(t)}, \mb{V}_{i}^{(t)}$, $\mb{w}_i^{(t-1)}$) considers the updated RAG (see Section \ref{sec:interaction}) at time $t$ and  computes the new weights of the GMM given the prior weights $w_i^{(t-1)}$.
    This update computes one $M$-step update of the EM algorithm for updating the weights in a GMM given the data $R_i^{(t)}$, means $\mb{M}_{i}^{(t)}$, covariances $\mb{V}_{i}^{(t)}$, and prior weights $\mb{w}^{(t-1)}_i$.
    \item[iii.] For integers $k,i\in [m]$, \texttt{GetKNearestGMMs}$\big(k, i ,t, \{\mb{w}_{j}^{(t)}\}_{j=1}^m\big)$ computes the agent indices of the $k$NN to agent $A_i^{(t)}$, where the distance between agents $A_i^{(t)}$ and $A_j^{(t)}$ is computed via $d(i,j)=\|\mb{w}_{i}^{(t)}-\mb{w}_{j}^{(t)}\|_2$.  Note that we adopt the convention that $i\notin \texttt{GetKNearestGMMs}\big(k, i ,t, \{\mb{w}_{j}^{(t)}\}_{j=1}^m\big)$.
\end{itemize}

\begin{algorithm}[t!]
\caption{Standard simulation procedure}
\label{alg:standard-simulation-procedure}
\begin{algorithmic}[1]
    \Procedure{SimulateInteractingGMMs}{$T, p, k, r, n, m,\{\mb{w}_i^{(0)}\}_{i=1}^m, \mb{M}, \mb{V}$, $\epsilon>0$}
    \For{$i \in \{1,\dots, m\}$}\Comment{Initialization}
    \State $R_i^{(0)} \gets $\texttt{SampleFromGMM}($\mb{w}_i^{(0)},\mb{M}, \mb{V}$, $r$)
    \EndFor
    \For{$t=1,2, \dots, T$} \Comment{Interaction}
    \For{$i \in \{1, \dots, m\}$}
        \State $u \gets $\texttt{Uniform}(0,1)
        \If{$u < p$}
            \State $j \gets i$
        \Else
            \State $\mathcal{N}_i\gets $\texttt{GetKNearestGMMs}$\big(k, i ,t, \{\mb{w}_{\ell}^{(t)}\}_{\ell=1}^m\big)$ 
            \State $j$ is uniformly chosen from $\mathcal{N}_i$
        \EndIf
        \State $x \gets$ \texttt{SampleFromGMM}($\mb{w}_i^{(t-1)},\mb{M},\mb{V}$, 1)\Comment{Query}
\State $\tilde R_j^{(t)}\gets (R_j^{(t)} \cup \{x\}) \setminus (\arg\max_{v \in R_j^{(t)}} \|x-v\|)$\Comment{Pseudo-update}
\State $\tilde {\mb{w}}_j^{(t)}\gets$ \texttt{UpdateGMM}$(\tilde R_j^{(t)},\mb{M},\mb{V},\mb{w}_j^{(t-1)}$)
\State y $\gets$ \texttt{SampleFromGMM}($\tilde {\mb{w}}_j^{(t)},\mb{M},\mb{V},1$)\Comment{Answer}
\State $R_{i}^{(t+1)} \gets (R_i^{(t)} \cup \{y\}) \setminus (\arg\max_{v \in R_i^{(t)}} \|y-v\|)$\Comment{RAG update}
\State $\mb{w}_i^{(t+1)} \gets$ \texttt{UpdateGMM}$(R_i^{(t+1)},\mb{M},\mb{V},\mb{w}_i^{(t)})$\Comment{Model update}
        \EndFor
    \EndFor
    \EndProcedure
\end{algorithmic}
\end{algorithm}
\noindent We now describe the three main phases of the interaction model; these are designed to mimic the interaction mechanism of the interacting LLM system in Section \ref{sec:LLM}.

\subsubsection{Initialization}
\label{sec:init}

The RAG initialization step corresponds to line 3 in Algorithm \ref{alg:standard-simulation-procedure}; in this step, we sample $r$ points from the GMM and set that as our starting RAG $R_i^{(0)}$ for the $i$th agent. 
This aims to model agents with diverse viewpoints (as each agent is certain of a different mean/topic) and unique knowledge and memories (i.e., unique RAGs) to justify their perspective. 
While Algorithm 1 accepts any set of initial weights \(\{\mb{w_i^{(0)}}\}_{i=1}^m\), for our experiments in Section \ref{sec:results}, we set $n=m=30$ unless otherwise specified and initialize the weight vector for the $i$th agent to be $\boldsymbol{w_i}^{(0)} = (1-\epsilon)\mb{\mathfrak{e}_i}+(\epsilon/n) \vec{1}_n\in\mathbb{R}^n$ where $\mathfrak{e}_i \in \mathbb{R}^n$ is the $i$th standard basis vector and $\vec{1}_n$ is the length $n$ vector of all 1's.
Note that $\epsilon>0$ here is meant to avoid degenerate weights in the update procedure, as in the EM update step weights of $0$ or $1$ will not change/be updated.


\subsubsection{Interaction} 
\label{sec:interaction}

For each of time $t= 1, \dots, T$, the GMM agents interact as described below.
\begin{itemize}
    \item[1.] \textbf{Choosing interaction pair:} At time $t>0$, each GMM agent $A_i^{(t)}$ (in sequence) chooses a GMM model $A_j^{(t)}$ to interact with as follows.
With probability $p$, agent $A_i^{(t)}$ chooses to self-interact (i.e., $A_j^{(t)}=A_i^{(t)}$); this mimics self-mirroring in the LLM system.
With probability $1-p$, agent $A_i^{(t)}$ chooses to interact with an agent $A_j^{(t)}$ uniformly selected from \texttt{GetKNearestGMMs}$\big(k, i ,t, \{\mb{w}_{\ell}^{(t)}\}_{\ell=1}^m\big)$.  This corresponds to lines 7 to 14 in Algorithm \ref{alg:standard-simulation-procedure}.
    \item[2.] \textbf{Question:} First, $A_i^{(t)}$ sends a query to $A_j^{(t)}$; this amounts to sampling $x$ from $\mathcal{F}_i^{(t)}$ and sending $x$ to $A_j^{(t)}$.
    This corresponds to line 15 in Algorithm \ref{alg:standard-simulation-procedure}.
    \item[3.] \textbf{Answer:} The $j$th agent $A_j^{(t)}$ responds to $x$ as follows.
Let $h=\text{argmax}_{\ell}\|x-x^{(t)}_{\ell j} \|_2$, and set $\tilde R_j^{(t)}=\{x^{(t)}_{1 j},\cdots,x^{(t)}_{(h-1) j},x,x^{(t)}_{(h+1) j},\cdots,x^{(t)}_{r j} \}$.  
Note that agent $A_j^{(t)}$ is not updated, this is more akin to a pseudo-update designed to factor the query $x$ into the response $y$.
We then set $\tilde w_j^{(t)}=\texttt{UpdateGMM}(\tilde R_j^{(t)},\mb{M},\mb{V},\mb{w}_j^{(t)}),$ and the response $y$ is generated via \texttt{SampleFromGMM}$(\tilde {\mb{w}}_j^{(t)},\mb{M},\mb{V},1)$, and $y$ is sent to $A_i^{(t)}$.
    This corresponds to lines 16-18 in Algorithm \ref{alg:standard-simulation-procedure}.
\item[4.] \textbf{Model Update:} The $i$th agent $A_i^{(t)}$ updates as follows.
Let $h=\text{argmax}_{\ell}\|y-x^{(t)}_{\ell i} \|_2$, and update the RAG of $A_i^{(t)}$ via
$R_i^{(t+1)}=\{x^{(t)}_{1 i},\cdots,x^{(t)}_{(h-1) i},y,x^{(t)}_{(h+1) i},\cdots,x^{(t)}_{r i} \}$.  
We then update $\mathcal{F}_i^{(t+1)}$ via  $w_i^{(t+1)}=\texttt{UpdateGMM}(R_i^{(t+1)},\mb{M},\mb{V},\mb{w}_i^{(t)}).$
    This corresponds to lines 19-20 in Algorithm \ref{alg:standard-simulation-procedure}.
\end{itemize}
After each agent is updated, we set the time index to $t+1$, and repeat the interaction process.
\begin{remark}
\label{rem:steps}
\emph{
In the above, one time step involves updating all $m$ GMM agents.  In the theory below, the process is slowed so that one time step involves updating a single random GMM agent.
The dynamics of the two updating systems are (up to a scaling factor) the same; the single agent update was used to help the theoretical tractability of the GMM framework.
}
\end{remark}
\begin{remark}
    \label{rem:sigma}\emph{
    If the covariance structure $\mathbf{V}$ is meant to capture the available information about a topic, then it is reasonable to keep  $\mathbf{V}$ fixed throughout the GMM interaction process.
    If $\mathbf{V}_i$ is meant rather to capture agent $i$'s notion of certainty, then it is reasonable to let  $\mathbf{V}_i$ vary in time throughout the process (then indexing it via $\mathbf{V}_i^{(t)}$).
    In this case, we can update the $\mathbf{V}_i^{(t)}$ again via one M-step of the EM algorithm; see Algorithm \ref{alg:variance-simulation-procedure} in Appendix \ref{app:exp_std} for detail.  
    In Appendix \ref{app:exp_std}, we consider interacting Gaussian agents (in 1-d) where we let both the weights and variances update through the process. 
    As suspected, these experiments reveal that the standard deviations decrease over time, eventually causing the Gaussian components to collapse into point-mass distributions. 
    In order to prevent this degeneracy, we propose a volume constraint update in which the determinant (volume) of the covariance matrices is fixed throughout the process (e.g., via rescaling them after each EM update iteration).
    This prevents singularities while allowing the covariances of the components to have variable orientation and shape, allowing agents to develop varying sensitivities towards different topic dimensions.
    This would more effectively model how an agent's focus may narrow in one direction while remaining broad in others without the model collapsing; see Section \ref{sec:conclusion} for example.
    }
\end{remark}

\subsection{The Markov chain induced by GMM interactions}
\label{sec:MC}
Our theoretical developments will be cast in the following Markov chain (MC) simplification of the the interacting GMM system above; for a general background on Markov chains, see \cite{douc2018markov,levin2017markov}.
For the sake of simplicity, our theory will be set in the setting of a mixture of two one-dimensional Gaussian distributions; note that extending to more Gaussian components in higher dimensions is relatively straightforward, see Remark \ref{rem:hd}.
We will also streamline the question/response protocol in Algorithm \ref{alg:standard-simulation-procedure} into a single update mechanism described below.

The state space of the Markov chain is
$$\bigg((0,1)\times \underbrace{\mathbb{R}\times\mathbb{R}\times\cdots \times\mathbb{R}}_{r\text{ times}}\bigg)^m,$$
where the state of the Markov chain at time $t$ being 
$\mathfrak{G}_t=\{(w_i^{(t)},x_{1i}^{(t)},\cdots,x_{ri}^{(t)})\}_{i=1}^m$ denotes that the $i$th GMM agent is a mixture model of the form 
$
\mathcal{F}^{(t)}_i=w_i^{(t)} \mathcal{N}(-1,\sigma^2)+(1-w_i^{(t)})\mathcal{N}(1,\sigma^2)
$ where $\mathcal{N}(\mu,\sigma^2)$ denotes the normal with mean $\mu$ and variance $\sigma^2$; moreover, the RAG for the $i$th GMM agent is $\{x_{1i}^{(t)},\cdots,x_{ri}^{(t)}\}$.
The Markov chain then evolves as follows.
\begin{itemize}
\item[1.]  One of the $m$ GMM's is chosen uniformly at random to be updated; denote the index of this model via $i$.
\item[2.]  One of the $m$ GMM's is chosen independently, uniformly at random to send the update; denote the index of this model via $j$. Note that $j$ could equal $i$, and this case provides a mirroring probability equal to $p=1/m$.
\item[3.] Independently generate $y\sim \mathcal{F}^{(t)}_j$. Letting $\ell=\text{argmin}_h \|y-x_{hi}^{(t)}\|_2$, update the RAG of the $i$th agent via 
$$
\{x_{1i}^{(t+1)},\cdots,x_{ri}^{(t+1)}\}=
\{x_{1i}^{(t)},\cdots,x_{(\ell-1) i}^{(t)},y,x_{(\ell+1)i}^{(t)},\cdots,x_{ri}^{(t)}\}.
$$
\item[4.] Update the weight of $\mathcal{F}^{(t)}_i$ via
$$
w_i^{(t+1)}=\frac{1}{r}\left(\sum_{h=1, h\neq\ell }^r h_\sigma(w_i^{(t)},x_{h i}^{(t)})+h_\sigma(w_i^{(t)},y)\right),
$$
where
$$h_\sigma(w,x)=\frac{w}{w+(1-w)e^{2x/\sigma^2}}=\frac{1}{1+\left(\frac{1-w}{w}\right)e^{2x/\sigma^2}}.$$
\end{itemize}
The update mechanism of this MC, adapted here to the 1-d case above, is the same as the update mechanism in Algorithm \ref{alg:standard-simulation-procedure}, with the core difference again being the streamlining of the question/response step.

\begin{figure}[t!]
    \centering
    \includegraphics[width=\linewidth]{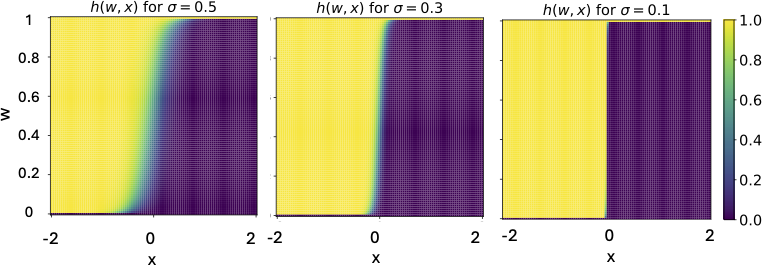}
    \caption{Heatmap visualization of the function $h(w, x)$ across three values of $\sigma$. 
    The function is plotted over the domains $w \in [0, 1]$ and $x \in [-2, 2]$. 
    The transition from the maximum value ($h \approx 1$, yellow) to the minimum ($h \approx 0$, dark purple) occurs at $x=0$. 
    As $\sigma$ decreases from $0.5$ to $0.1$, the gradient sharpens significantly, causing the function to approach a step-function behavior. } 
    \label{fig:h_function}
\end{figure}

\begin{remark}
    \emph{For \(w\in(0,1)\), $\lim_{x\rightarrow \infty}h_\sigma =0$  and $\lim_{x\rightarrow -\infty}h_\sigma =1$. As illustrated in Figure \ref{fig:h_function}, the parameter \(\sigma\) dictates the sharpness of this transition. For small \(\sigma\), the function \(h_\sigma\) behaves close to a step function, but for large \(\sigma\), the transition is smoother, allowing \(h_\sigma\) to take intermediate values in \((0,1)\), which increases the probability of moving to other silos for values of $x$ that are closer to 0.
}
\end{remark}

\begin{remark}
\label{rem:hd}
    \emph{Much of our work translates easily to the setting of higher-dimensional Gaussian mixture components under mild model assumptions.
    To wit, consider $n\leq 2^d$ Gaussian components in $\mathbb{R}^d$, where the mean vectors are all in $\{-1,1\}^d$.  
    If the $i$-th component has covariance $\Sigma_i=\sigma^2 I_d$ (where $I_d$ is the d-dimensional identity matrix), then the weight updates for $i$th model with RAG $
\{x_{1i}^{(t+1)},\cdots,x_{ri}^{(t+1)}\}$ and weight $\mb{w}^{(t)}_i$ are given by
$$
(\mb{w}_i^{(t+1)})_j=\frac{1}{r}\left(\sum_{h=1}^r g_{j,\sigma}(\mb{w}_i^{(t)},x_{h i}^{(t)})\right),
$$
where 
\begin{align*}
g_{j,\sigma}(\vec{w},\vec{x})&=\frac{ w_j e^{\sum_{\ell=1}^d x_\ell \mu_{j\ell}/\sigma^2 }
}{
\sum_{h=1}^n w_h e^{\sum_{\ell=1}^d x_\ell \mu_{h\ell}/\sigma^2 }}\\
&=\frac{ w_j 
}{
w_j+\sum_{h=1,h\neq j}^n w_h e^{\vec{x}^T(\vec{\mu}_h-\vec{\mu}_j)/\sigma^2 }}
\end{align*}
which is similar to the function $h_{\sigma}$ utilized in the one-dimensional setting.
See Remark \ref{rem:hdpol} in Section \ref{sec:MCPol} for discussion on how to show polarization in this high dimensional setting.
}
\end{remark}

\section{Silos and polarization}
\label{sec:silo}

At each time $t = 0, 1, \dots, T$, we evaluate each model to determine which of the $n$ perspectives it is most aligned with; this amounts to determining the mixture component with the highest mixing weight in the GMM model (breaking ties by picking the smaller index). 
 We call this the \emph{silo} to which the agent belongs, and define 

\begin{equation}
    \operatorname{silo}\Big(A_i^{(t)}\Big) = \argmax_{j \in \{1, \dots, n\}} \bigg(\Big(\mb{w}_i^{(t)}\Big)_{j}\bigg)
\end{equation}


\noindent The silo of an agent represents a (relatively weak) notion of attention for the agent, since agents that belong to the same silo have similar beliefs and are considered aligned. 
\emph{Polarization} is a form of extreme siloing, where if $\operatorname{silo}\big(A_i^{(t)}\big)=\ell$, then $(\mb{w}_i^{(t)})_\ell$ is sufficiently close to 1.
We next define this notion rigorously and explore it theoretically in the context of the Markov chain in Section \ref{sec:MC}.

\subsection{Polarization in the Markov chain}
\label{sec:MCPol}
In the MC model of Section \ref{sec:MC}, it is sensible to say that the system is polarized if all agents are sufficiently focused on one of the Gaussian mixture components; i.e., if their associated weight $w$ is sufficiently close to $0$ or $1$.
To quantify this, we first define the parameter $\rho<1/2$ and define
\begin{align*}
    \eta&=\eta_{\rho,\sigma}=\mathbb{P}(\mathcal{N}(-1,\sigma^2)\in -1\pm \rho); \quad C=C_{\rho,\sigma}=e^{2(1-\rho)/\sigma^2}.
\end{align*}
This $\rho$ is chosen so that the RAG behaves as follows during the update step of the MC: 
Assume that all existing RAG elements are in $(1\pm \rho)\cup(-1\pm \rho)$ and an update in $1\pm \rho$ is received (with the $-1\pm \rho$ case following analogously). 
The element removed from the RAG can only be in $1\pm \rho$ if there are no RAG elements in  $-1\pm \rho$.

We next define the notion and degree of system polarization.
For each $\ell\in\mathbb{Z}>0$, define the intervals $\mathcal{I}_\ell=\mathcal{I}_{r,\sigma,\ell}=\left(\frac{1}{1+C^\ell},\frac{1}{1+C^{-\ell}} \right)$.
For integers $\ell>0$ and $t>0$, define the event $E^{(t)}_\ell$ to be the event that 
\begin{itemize}
    \item[i.] At time $t>0$, all weights $\{w_i^{(t)}\}_{i=1}^m$ are \emph{outside} the interval $\mathcal{I}_\ell$;
    \item[ii.] If for an index $i$, we have that $w_i^{(t)}<\mathcal{I}_\ell$, then all $\{x_{j i}\}_{j=1}^r $ are in  $1\pm \rho$;
    \item[iii.] If for an index $i$, we have that $w_i^{(t)}>\mathcal{I}_\ell$, then all $\{x_{j i}\}_{j=1}^r $ are in  $-1\pm \rho$.
\end{itemize}
\begin{defn}
    If the configuration of the GMM MC at time $t>0$ is given by 
    $$\mathfrak{G}_t=\{(w_i^{(t)},x_{1i}^{(t)},\cdots,x_{ri}^{(t)})\}_{i=1}^m,$$ then we say this configuration is \emph{level-$\ell$ polarized} if $E^{(t)}_\ell$ holds.
\end{defn}
Our first main theoretical result, Theorem \ref{thm:polar}, gives a uniform lower bound on the probability of the MC system polarizing after a fixed number of steps.
As the system is Markov, this ensures that polarization occurs almost surely as $t\rightarrow \infty$.
\begin{thm}
\label{thm:polar}
There is a constant $q=q_{m,\rho,r}>0$ such that 
    $$
    \mathbb{P}\left(E_1^{(t+(5r+2)m)}\,\big|\,\text{state of GMM MC at time }t\text { is } \{(w_i^{(t)},x_{1i}^{(t)},\cdots,x_{ri}^{(t)})\}_{i=1}^m\right)\geq q
    $$
    holds for any $\{(w_i^{(t)},x_{1i}^{(t)},\cdots,x_{ri}^{(t)})\}_{i=1}^m$ in the GMM MC state space.
\end{thm}
The proof of Theorem \ref{thm:polar} can be found in Appendix \ref{app:proof1}.

\begin{remark}
    \label{rem:hdpol}\emph{
    In the high dimensional case outlined in Remark \ref{rem:hd}, note that if $h\neq j,$ then $\|\vec{\mu}_j-\vec{\mu}_h\|\geq 2.$
If $\vec{x}$ is in a ball of radius $r\leq 1/2$ about $\vec{\mu}_j$, then noting $0\leq \|\vec{\mu}_j-\vec{x}\|^2\leq r^2$ and 
    $(2-r)^2\leq \|\vec{\mu}_h-\vec{x}\|^2\leq (2+r)^2,$
we have that (as all $\vec{\mu}_i$ have the same norm)
\begin{align*}
    x^T(\vec{\mu}_h-\vec{\mu}_j)&=(\|\vec{\mu}_j-x\|^2-\|\vec{\mu}_h-x\|^2)/2\leq (r^2-(2-r)^2)/2=-2+2r\\
    x^T(\vec{\mu}_h-\vec{\mu}_j)&=(\|\vec{\mu}_j-x\|^2-\|\vec{\mu}_h-x\|^2)/2\geq -(2+r)^2/2.
\end{align*}
Therefore, if $\vec{x}$ is in a ball of radius $r\leq 1/2$ about $\vec{\mu}_j$ then
\begin{align*}
g_{j,\sigma}(\vec{w},\vec{x})&\leq\frac{ w_j 
}{
w_j+\sum_{h=1,h\neq j}^m w_h e^{-(2+r)^2
/(2\sigma^2) }}\\
g_{j,\sigma}(\vec{w},\vec{x})&\geq \frac{ w_j 
}{
w_j+\sum_{h=1,h\neq j}^m w_h e^{(2r-2)
/\sigma^2 }}
\end{align*}
and hence if all RAG elements are in a ball of radius $r\leq 1/2$ about $\vec{\mu}_j$, then
\begin{align*}
(\mb{w}_i^{(t+1)})_j&\leq\frac{ w_j 
}{
w_j+\sum_{h=1,h\neq j}^m w_h e^{-(2+r)^2
/(2\sigma^2) }};\\
(\mb{w}_i^{(t+1)})_j&\geq\frac{ w_j 
}{
w_j+\sum_{h=1,h\neq j}^m w_h e^{(2r-2)
/\sigma^2 }}.
\end{align*}
These bounds mimic those in the proof of Theorem \ref{thm:polar}, and would allow for a straightforward (though notationally unwieldy) adaptation to the proof in the 1-d case; details are omitted for brevity.
}
\end{remark}

\subsection{Polarization stability}
\label{sec:stable}
In \cite{mcguinness2024investigating}, the authors conduct an extensive study of the effect of the global interaction parameters $k$ and $p$ on siloing behavior, where $k$ represents the number of closest neighbors an agent can talk to and $p$ represents the mirroring probability or probability an agent interacts with itself.
While we explore this empirically below, herein we explore this stability theoretically.
We do note here that as the GMM system updates are Gaussian (and hence unbounded), there is non-zero chance of getting arbitrarily extreme responses to queries in the GMM system.  
Given enough extreme responses in sequence, as the weight updates are controlled via the M-step of EM, an agent can have its weight vector pushed from any point in $(0,1)$ to any other point in $(0,1)$, and therefore that the polarization will not be globally stable into the infinite time horizon; this is related to the concept of \emph{metastability} in the stochastic process literature \citep{koralov2024metastable}.

Consider the setting where the MC is level-$\ell$ polarized at time $t$ and that we have a $k$-restriction on the communication mechanism (i.e., each GMM can only communicate with its $k$ nearest neighbors).  
Let $\sigma^2$ and $\rho$ be such that $C>2$, and assume that 
$$
\min\left(|i:w_i^{(t)}<\mathcal{I}_{\ell}| ,|i:w_i^{(t)}>\mathcal{I}_{\ell}|\right)>k.
$$
After $s$ steps the MC is still level-$\ell$ polarized with probability at least (as $1+C^{-\ell}<e^{C^{-\ell}}$)
$$
\left(\frac{\eta}{1+C^{-\ell}}\right)^{s}>\eta^se^{-sC^{-\ell}}.
$$
Moreover, if each GMM is selected at least once and all updates for the $w_i<\mathcal{I}_\ell$ come from $1\pm \rho$ and for all $w_i>\mathcal{I}_\ell$ from $-1\pm \rho$, then the GMM would be level-$(\ell+1)$ polarized.
Thus we have the following, where the final bound follows from simple coupon collector asymptotics.
\begin{table}[t!]
\begin{tabular}{c|ccccc}
$\sigma\backslash\ell$ & 1 & 2 & 3 & 4 & 5\\ \hline
0.3 & 2.984834e-45 & 3.030700e-45 & 3.030701e-45 & 3.030701e-45 & 3.030701e-45\\
0.2 &  2.897895e-06 & 2.897895e-06 & 2.897895e-06 & 2.897895e-06 & 2.897895e-06\\
0.1 &  9.994152e-01 & 9.994152e-01 & 9.994152e-01 & 9.994152e-01 & 9.994152e-01\\
0.05&  1.000000e+00 & 1.000000e+00 & 1.000000e+00 & 1.000000e+00 & 1.000000e+00
\end{tabular}
\caption{Lower bounds in Theorem \ref{thm:stable} in the setting where $c=10$, $m=30$ and $\rho=1/2$; note that the bound in part ii. of the Theorem is identical (up to the precision in the table) to the bound in part i. of the Theorem.}
\label{tab:tab2}
\end{table}

\begin{thm}
\label{thm:stable}
    With notation and assumptions as above, if the GMM MC at time $t$ is level-$\ell$ polarized with 
$$
\min\left(|i:w_i^{(t)}<\mathcal{I}_{\ell}| ,|i:w_i^{(t)}>\mathcal{I}_{\ell}|\right)>k,
$$ 
then after $s= cm\log m$ steps the GMM MC with $k$-restricted communication will 
\begin{itemize}
    \item[i.] remain level-$\ell$ polarized with probability at least $\eta^{cm\log m}\text{exp}\left\{-\frac{cm\log m}{C^\ell}\right\}$;
    \item[ii.] will become level-$(\ell+1)$ polarized with probability at least $$\eta^{cm\log m}\text{exp}\left\{-\frac{cm\log m}{C^\ell}\right\}(1-m^{-c+1}).$$
\end{itemize}
\end{thm}

The lower bounds in Theorem \ref{thm:stable} are a bit opaque, so to explore them further we consider the lower bounds in the setting where $c\in\{2,10\}$, $m=30$ and $\rho=1/2$; the values of the lower bounds---as a function of $\ell$ and $\sigma$---are given in Tables \ref{tab:tab2}--\ref{tab:tab4}.
Note that after $cm\log m$ steps in the MC, standard coupon collector asymptotics give that with high probability (for large enough $m$) each agent has been updated at least $c$ times; this would then correspond to roughly $c$ steps being taken in the interacting GMM system defined by Algorithm \ref{alg:standard-simulation-procedure}.
When $c$ is small, we see that for modest $\sigma$ (here $\sigma\leq 0.2$) there is a high probability of remaining polarized after $cm\log m=60\log 30\approx 200$ steps.  In the setting where $c\geq 10$ (so that the system has taken $\approx 1000$ steps), a smaller $\sigma$ is needed for high-probability stable poles, though $\sigma\leq 0.1$ seems to be sufficient in both cases.

\begin{table}[t!]
\begin{tabular}{c|ccccc}
$\sigma\backslash\ell$ & 1 & 2 & 3 & 4 & 5\\ \hline
0.4  & 6.188271e-22 & 9.169169e-22 & 9.176131e-22 & 9.176144e-22 & 9.176144e-22\\
0.3  & 1.244469e-09 & 1.248270e-09 & 1.248270e-09 & 1.248270e-09 & 1.248270e-09\\
0.2  & 7.805784e-02 & 7.805784e-02 & 7.805784e-02 & 7.805784e-02 & 7.805784e-02\\
0.1  & 9.998830e-01 & 9.998830e-01 & 9.998830e-01 & 9.998830e-01 & 9.998830e-01\\
0.05 & 1.000000e+00 & 1.000000e+00 & 1.000000e+00 & 1.000000e+00 & 1.000000e+00
\end{tabular}
\caption{Lower bounds in Theorem \ref{thm:stable} part i. in the setting where $c=2$, $m=30$ and $\rho=1/2$.}
\label{tab:tab3}
\begin{tabular}{c|ccccc}
$\sigma\backslash\ell$ & 1 & 2 & 3 & 4 & 5\\ \hline
0.4  & 5.981996e-22 & 8.863530e-22 & 8.870260e-22 & 8.870273e-22 & 8.870273e-22\\
0.3  & 1.202987e-09 & 1.206661e-09 & 1.206661e-09 & 1.206661e-09 & 1.206661e-09\\
0.2  & 7.545591e-02 & 7.545591e-02 & 7.545591e-02 & 7.545591e-02 & 7.545591e-02\\
0.1  & 9.665536e-01 & 9.665536e-01 & 9.665536e-01 & 9.665536e-01 & 9.665536e-01\\
0.05 & 9.666667e-01 & 9.666667e-01 & 9.666667e-01 & 9.666667e-01 & 9.666667e-01
\end{tabular}
\caption{Lower bounds in Theorem \ref{thm:stable} part ii. in the setting where $c=2$, $m=30$ and $\rho=1/2$.}
\label{tab:tab4}
\end{table}

\section{Empirical results}
\label{sec:results}

Herein, we present results exploring the formation of dynamic stable/non-stable silos in the interacting GMM system of Algorithm \ref{alg:standard-simulation-procedure}. The code can be found at \url{https://github.com/edlwang/GMM-LLM-Proxy}.

\subsection{Silo Patterns}
\label{ssec:unstable-silos}

In \cite{mcguinness2024investigating}, three distinct siloification behaviors are considered:stable silos, unstable silos, and decaying silos.
We focus on the cases of stable and unstable silos, defined by the following stability metric.
    The \emph{stability} of the interacting GMM system at time $t$---denoted $S^{(t)}$---is defined to be the proportion of agents that have changed silos from time step $t-1$ to $t$.  To wit, $S^{(t)}$ is defined via
\begin{equation}
\label{eq:stable}
    S^{(t)} = \dfrac{1}{n}\sum_{i=1}^n 1_{\operatorname{silo}(A_i^{(t)}) \neq \operatorname{silo}(A_i^{(t-1)})}
\end{equation}
We then define stable and unstable silos below.
\begin{defn}
  For an integer $T>0$, we say that in the interacting system of GMMs a \emph{length $T$ stable silo system occurs at time $t_0$} if 
  \begin{itemize}
      \item[i.] $S^{(t)} = 1$ for all $t\in[t_0,t_0+T]$.
      \item[ii.] $\left|\left\{\operatorname{silo}\Big(A_i^{(t)}\Big)\right\}_{i=1}^m \right|$ is constant for all $t\in[t_0,t_0+T]$.
  \end{itemize}
    For an integer $T>0$, we say that in the interacting system of GMMs a \emph{length $T$ unstable silo system occurs at time $t_0$} if 
  \begin{itemize}
      \item[i.] $S^{(t)} < 1$ for some $t\in[t_0,t_0+T]$.
      \item[ii.] $\left|\left\{\operatorname{silo}\Big(A_i^{(t)}\Big)\right\}_{i=1}^m \right|$ is constant for all $t\in[t_0,t_0+T]$.
  \end{itemize} 
\end{defn}

\begin{figure}[t!]
    \centering
    \includegraphics[width=1\textwidth]{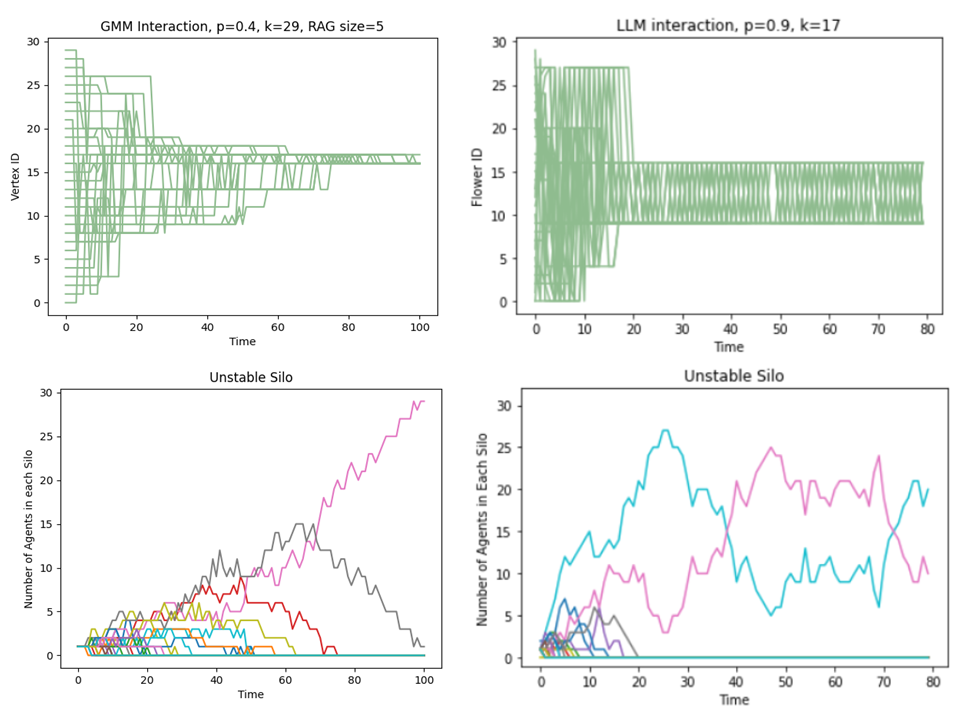}
    \caption{Comparison of unstable silo behavior for our GMM simulation and the LLM simulation from \cite{mcguinness2024investigating} \textbf{Left.} Unstable silo behavior for the GMM model described in this paper. We set the global parameters to be $p=0.4$, $k=29$, $r=5$, and $T=100$. \textbf{Right.} Unstable silo behavior for the LLM model in \cite{mcguinness2024investigating} \textbf{Top.} An example unstable silo system where each line represents an agent. 
    \textbf{Bottom.} The evolution of the number of agents in each possible silo where each line represents a silo.
    }
    \label{fig:unstable-silo}
\end{figure}

\noindent In \cite{mcguinness2024investigating}, the authors explored the formation of stable/unstable silos in the interacting LLM system for varying system parameters (e.g., $r$, $k$, $p$, etc.).  Our goal is to mirror these phenomena in our system of interacting GMMs.

We first run our GMM system (here in 1-d) with RAG size $r = 5$, mirroring probability $p = 0.4$, and no neighborhood restriction (i.e., $k=29$).
We consider $t=0,1,\cdots,100$ steps, and plot the results of our GMM system (Left panels) and the motivating LLM system (Right panels) in Figure \ref{fig:unstable-silo}.
The first column shows our results using GMMs and the second column shows the results from figure 2 in McGuinness et al. using LLMs. 
In both columns, we see unstable silos forming by time 100.  
In the top panels, we plot the silo membership for each of the 30 agents, where each line represents an agent. 
In both setups, we see that after a certain time $t$, approximately 75 for the GMM case and 22 for the LLM case, only two silos exist for the rest of the simulation. 
Furthermore, we see lines jumping between the two silos, indicating that these silos are unstable. 
Moreover, we see that in our GMM case it takes significantly more time for the unstable silos to form and that the switching between the two silos is slower compared to the LLM case.


The second row of figures represents the evolution of the number of agents in each possible silo with each line representing a particular silo. In both cases, we see that the number of agents in all silos, except for two, go to zero.
In the GMM setting, after two silos emerge and briefly oscillate, one begins to slowly dominate leading to convergence to a single silo.
This contrasts the LLM case, where number of agents in the remaining two silos oscillate (though in longer time horizons, we see the LLMs converge to a single silo as well), and these silos do not have substantial decay or growth. 
We investigate this phenomenon of convergence to a single silo as $t\to\infty$ further in section \ref{subsec:single-silo}.
We again see that in the GMM system we see a longer time until formation of unstable silos and a slower speed of oscillation.


\subsection{Effect of neighborhood interaction parameter on silos}
\label{ssec:nneighbors}

\begin{figure}[t!]
    \centering
    \includegraphics[width=1\linewidth]{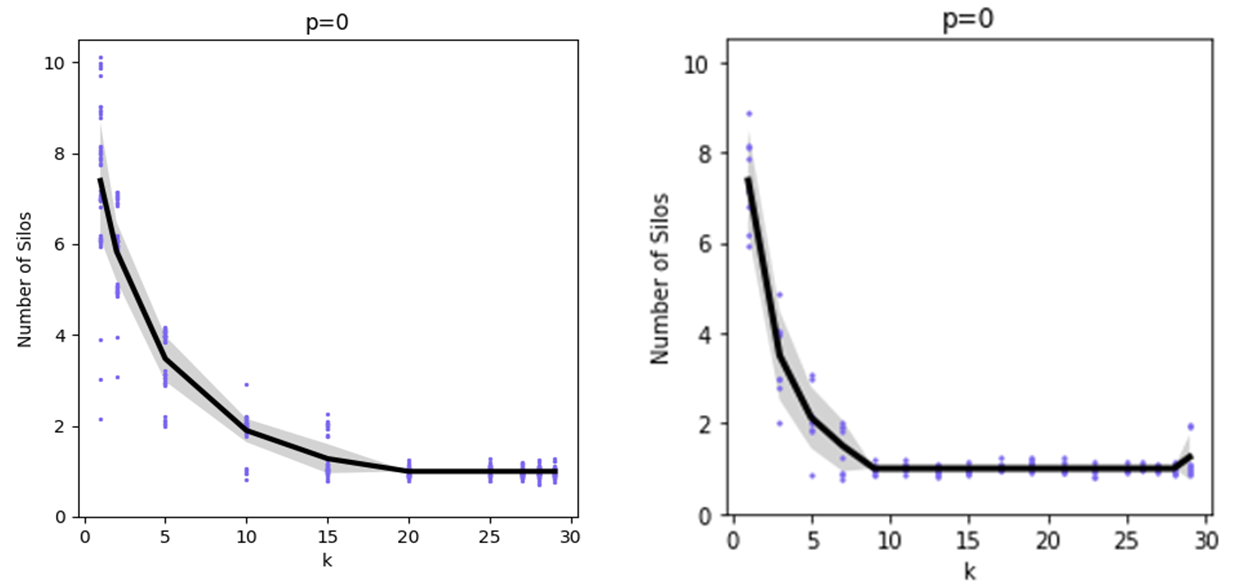}
    \caption{Comparison of the effect of $k$ on the number of silos for $p=0$ between the GMM and LLM simulations. 
    \textbf{Left.} The result of the GMM simulation with $T=80$, and $r=5$. Each value of $k$ was simulated 50 times with the line indicating the average and the shaded region indicating +/- 5 SE. \textbf{Right.} The result of the LLM simulation from Figure 4 of \cite{mcguinness2024investigating}}
    \label{fig:silosvsk}
\end{figure}

\cite{mcguinness2024investigating} conduct an extensive study of the effect of the global interaction parameters $k$ and $p$ on siloing behavior, 
where $k$ represents the number of closest neighbors an agent can talk to and $p$ represents the mirroring probability or probability an agent interacts with itself. 
In one experiment, they investigate how varying $k$ affects the number of silos formed. 
We focus on replicating these results for $p=0$ while varying $k$. 
For each value of $k$, we run 50 replicates of the simulation (again using 1-d GMM's) with $p=0$, $t=1,2,\cdots,80$, and $r=5$. 
Figure \ref{fig:silosvsk} compares our results on the left to the results from \cite{mcguinness2024investigating} on the right. 
We see that our simulations agree with the LLM-system findings of an inverse relationship where increasing $k$ and the ability to communicate globally results in the formation of fewer silos.

\begin{figure}[t!]
        \centering
        \includegraphics[width=1\textwidth]{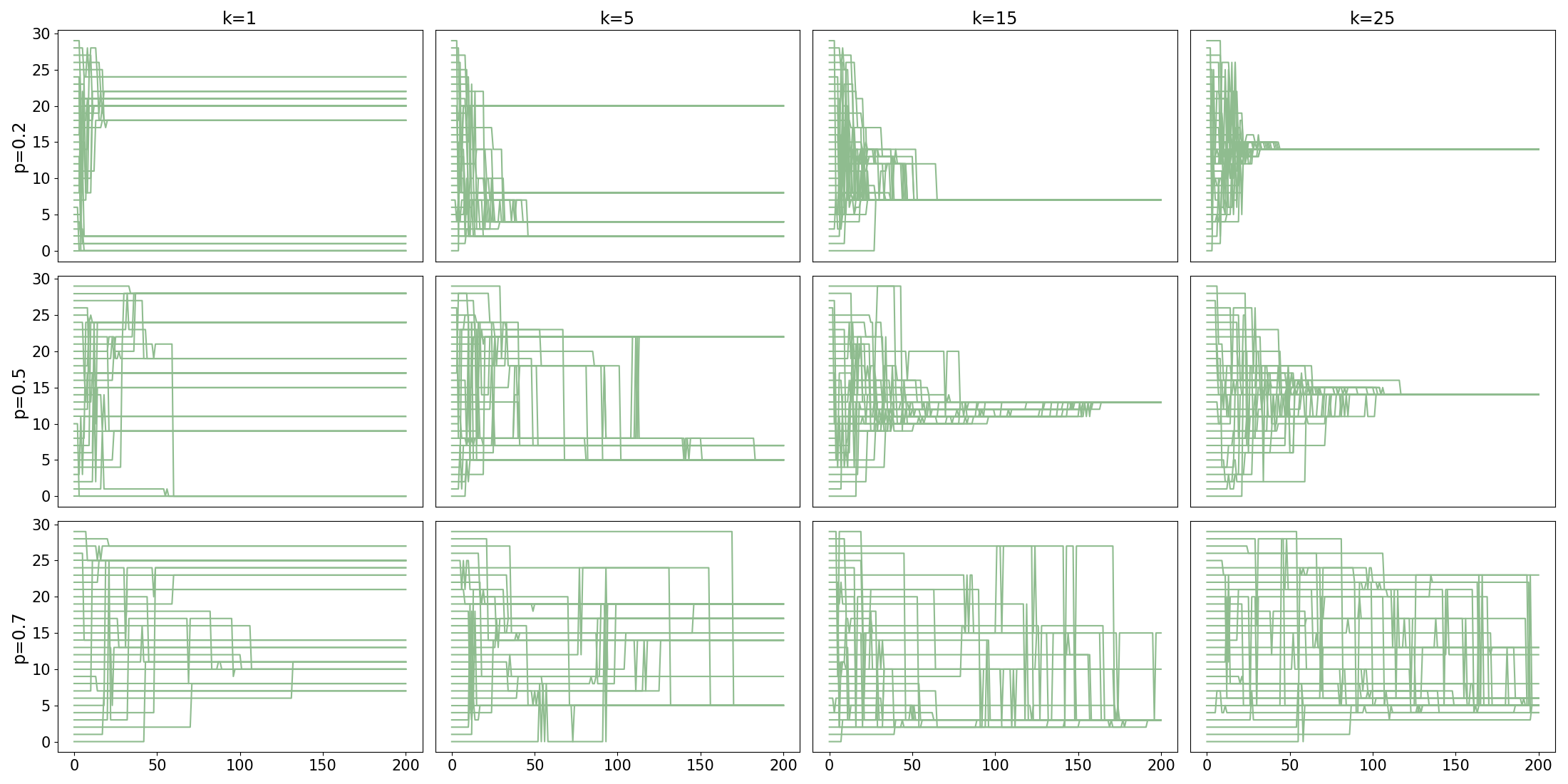}
        \caption{\textbf{GMM System.} Each plot has the time from $t = 0$ to $200$ on the x-axis and the silo of the agent on the y-axis.
        We plot example systems of $n=30$ interacting agents for varying values of $p$ and $k$ the GMM interaction system. 
        The value of $p$ is constant for each row and the value of $k$ is constant for each column.}
        \label{fig:gmmsweep}
        \end{figure}

\begin{figure}[t!]
        \includegraphics[width=1\textwidth]{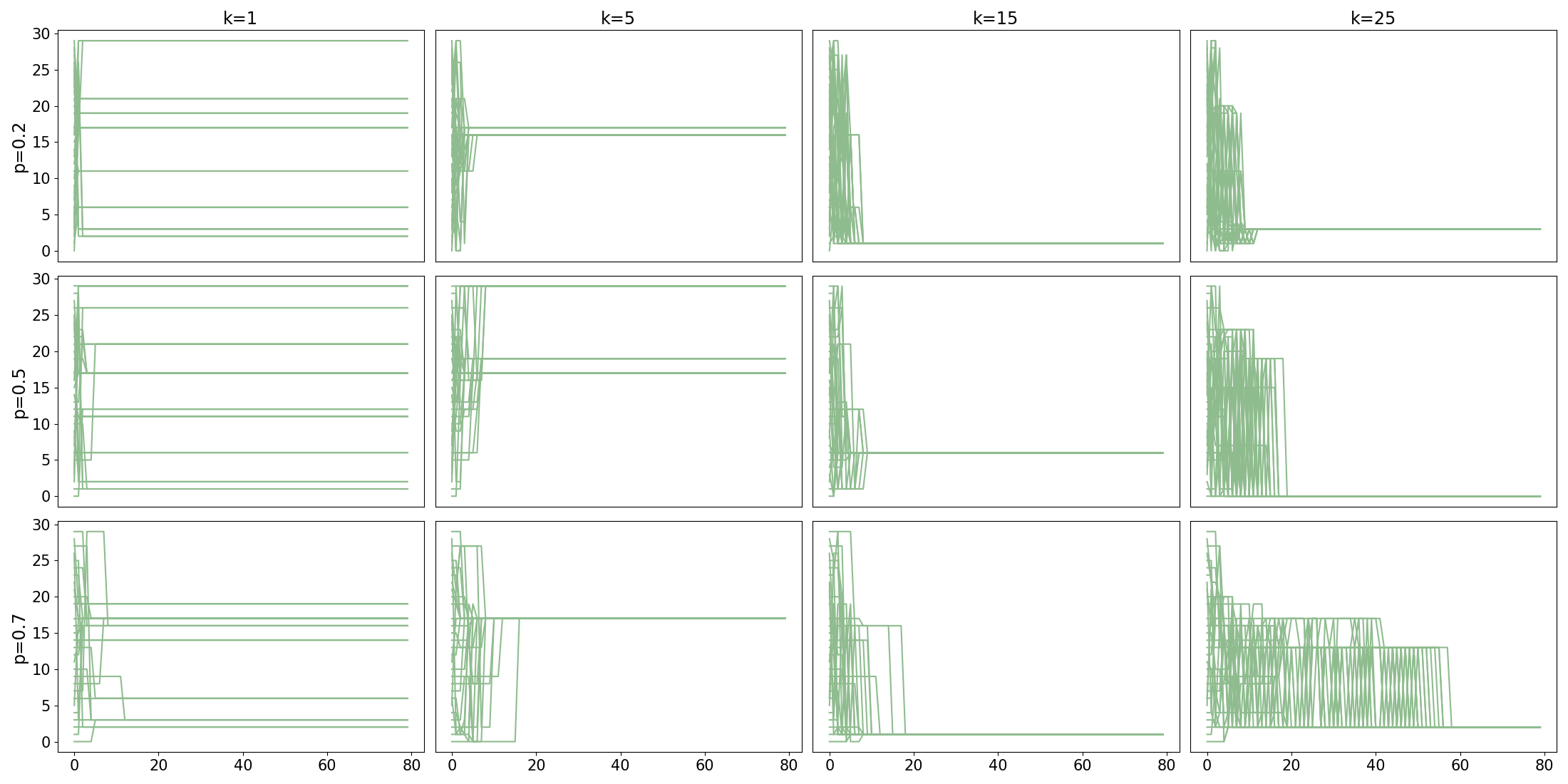}
        \caption{\textbf{LLM System.} Each plot has the time from $t = 0$ to $80$ on the x-axis and the silo of the agent on the y-axis. We plot example systems of $n=30$ interacting agents for varying values of $p$ and $k$ the LLM interaction system. 
        The value of $p$ is constant for each row and the value of $k$ is constant for each column.}
        \label{fig:llmsweep}
\end{figure}

Additionally, \cite{mcguinness2024investigating} provides a qualitative analysis of how agent behavior changes over a sweep of values of both $p$ and $k$. 
We demonstrate similar dynamics with GMM simulations (in 1-dimension) in Figure \ref{fig:gmmsweep}, which shows the interacting system dynamics for the GMM case (similar to the top panels in Figure \ref{fig:unstable-silo}, where each line represents an agent, and we plot the silo membership of each agent in time).
Figure \ref{fig:llmsweep} shows the LLM case. 
In both scenarios, we see that increasing $p$ slows convergence to stable or unstable silos, resulting in a longer time to achieve global alignment. In particular, the last row of Figure \ref{fig:gmmsweep} when $p=0.7$ shows behavior where the system has not converged to either a stable or an unstable silo yet. We call this a ``decaying silo," which is not a stable state but we expect that under more iterations that it will converge to either the stable or unstable silo case. Increasing $k$ decreases the number of silos formed, demonstrating that opportunities to communicate outside of a small neighbor set (mimicking talking to increasingly diverse people in the social network motivation) promotes global alignment.  

\begin{figure}[t!]
    \centering
    \includegraphics[width=1\linewidth]{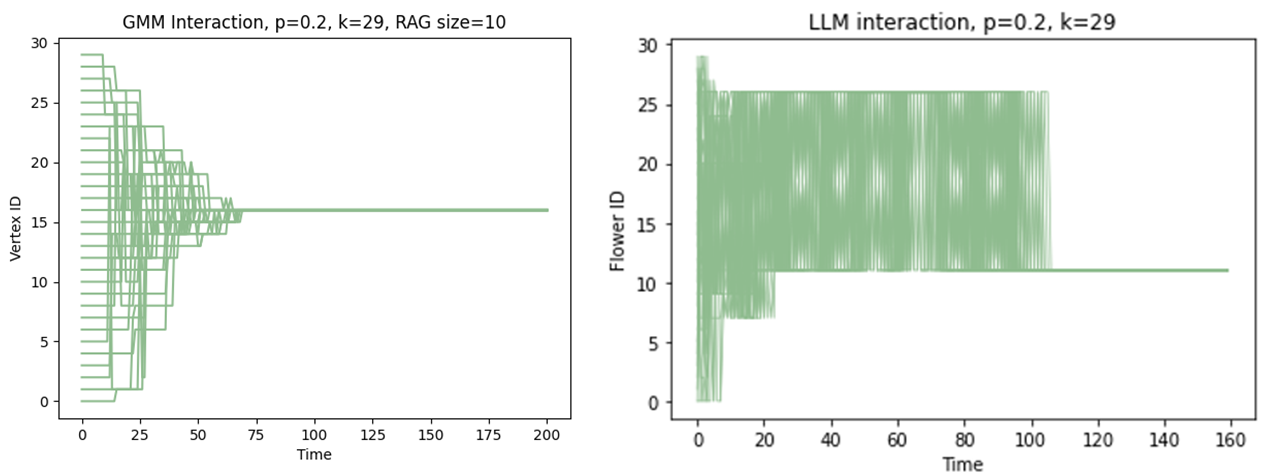}
    \caption{Example plot of the convergences to one silo for large $t$. Each line represents an agent. \textbf{Left.} The result of the GMM simulation with $p=0.2$, $k=29$, $T=200$, and $r=10$. \textbf{Right.} The result of the LLM simulation with $p=0.2$, $k=29$, $T=160$.}
    \label{fig:singlesilo}
\end{figure}

\subsection{Formation of a single silo for large t}
\label{subsec:single-silo}

When comparing the two unstable silo examples for the GMM and LLM case, 
we see that in the GMM system
it appears as though one length $T$ stable silo forms, where $T$ is relatively large.
This does not seem to occur in the LLM system in Figure \ref{fig:unstable-silo}, as it is noted in \cite{mcguinness2024investigating} there is no long-term pattern of growth or decay in the two silos. 
This is likely due to the small number of time steps for which the LLM simulation was run; indeed, in Figure \ref{fig:singlesilo} we demonstrate that both systems degenerate into a single stable silo given a long enough time horizon in the setting where $p=0.2$, $k=29$, $T=200$, and $r=10$ for 50 replicates. In every single experiment, we see a collapse into a single stable silo by $t=200$ in the GMM system and a similar collapse in the LLM system.
As one silo begins to dominate, we expect to see agents begin to choose neighbors to update with in that dominating silo, resulting in a positive feedback loop where other silos get absorbed into the dominant silo.

\begin{figure}
    \centering
    \includegraphics[width=\linewidth]{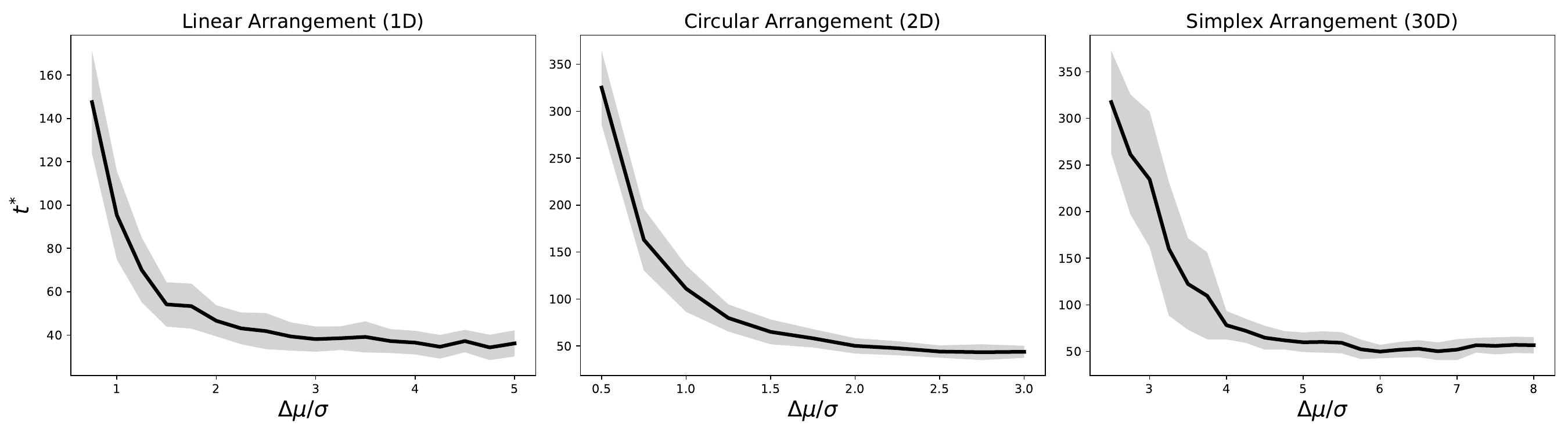}
    \caption{Convergence time \(t^{*}\) as a function of component separation across three geometric arrangements. \textbf{(Left)} Linear arrangement in one-dimensional space. \textbf{(Middle)} Circular arrangement in two-dimensional space. \textbf{(Right)} Simplex arrangement in 30-dimensional space. In all cases, the solid black line represents the mean convergence time across 50 replicates, while the shaded region indicates the \(\pm5\) standard error (SE) interval. A consistent trend is observed where \(t^{*}\) decreases as separation increases before reaching a plateau where the Gaussian components have negligible overlap and effectively disjoint supports.}
    \label{fig:deltamu}
\end{figure}

We also investigate the relationship that changing $\mb{M}$ (the means of the Gaussian mixture components) has on the convergence to a single silo in three different settings. In the first setting, we use the same linear initialization as our previous experiments in one-dimensional space. 
We let $\Delta \mu$ range from $0.75\sigma$ to $5\sigma$ with a step size of $0.25\sigma$ and set $\mu_i = \Delta\mu \cdot i$, where increasing $\Delta\mu$ increases the distances between the means of the Gaussians. Fixing $k=29$, $p=0$, and $r=5$, we ran the simulation 50 times for each value of $\Delta \mu$. In each simulation, we recorded $t^*$, the smallest time where for all $t > t^*$, the number of silos at time $t$ was $1$. The left plot of Figure \ref{fig:deltamu} shows that the average time it took to converge to a single silo decreased as we increased the distance between means. In our second setting, we moved to a more symmetrical configuration by putting our Gaussian mixture components in a two-dimensional space on a circle of radius $r$ around the origin. Therefore, we have \(\mb\mu_i = (r\cos(i\cdot 2\pi/30), r \sin (i\cdot 2\pi/30))\). The covariance matrix in this setting is \(\mb \Sigma = \sigma^2\mb {I_2}\). In this configuration, \(\Delta\mu\), the closest distance between two is given by \(\Delta\mu = 2r\sin(\pi/30)\). We let \(\Delta\mu\) to range from \(0.5\sigma\) to \(3\sigma\) with steps of \(0.25 \sigma\). Consistent with our first experiment, we performed 50 replicates for each \(\Delta\mu\), while keeping all other parameters identical. The result can be found in the middle plot of Figure \ref{fig:deltamu}. In our last experiment, we tried the most symmetrical arrangement of the component means by placing them in a 30-dimensional space on scaled unit vectors, which causes them to be equidistant from one another. In this case, we also used the symmetrical covariance matrix \(\mb \Sigma = \sigma^2\mb{I_{30}}\) and we let \(\Delta \mu\), the distance between the components to range from \(2.25\sigma\) to \(6\sigma\) with steps of \(0.25\sigma\); this is equivalent to \(\mb{\mu_i} = (\Delta\mu/\sqrt{2}) \,\mb{e_i}\), where \(\mb{e_i}\) is the standard basis vector for the i-th dimension. The result for this experiment can be found in the right plot of Figure \ref{fig:deltamu}. Note that in all three experiments the pattern is the same: the convergence time decreases as the distance increases, however, after some point, increasing the distance has no further effect on the time, as the components are already so far from each other that they have minimal overlap on one another.

\section{Discussion}
\label{sec:discussion}

In Sections \ref{sec:results} we compare the results of systems of interacting GMMs with that of interacting LLMs along numerous experimental conditions.
Section \ref{ssec:unstable-silos} focuses on the behavior of unstable silos, while section \ref{ssec:nneighbors} investigates the effect of the global parameters $p$, the probability of mirroring, and $k$, the number of nearest neighbors that each agent can communicate with, on the number of silos formed. 

In general, our results show many similarities between the behavior of systems of interacting GMMs and systems of interacting LLMs.
However, two main differences show up between the LLM and GMM behavior. Firstly, the system of interacting GMMs shows slower siloing and slower oscillatory behavior while LLM shows faster oscillatory behavior. 
This is likely the result of the difference in the incorporation of new information between the two models as well as inherent instability in the measurement step used for the LLM model. 
In the LLM model, updates are added to the RAG and the RAG is used when generating responses to the questions.
Thus, the interaction step has an immediate effect on the agent's response in the following time steps.
In the GMM, the response is added to the RAG which used to update the model parameters. 
However, the RAG is not directly used when generating responses.
This is a subtle point, and we suspect that the effect is that 
at each time step, each GMM agent is not changed nearly as much as the LLMs analogues change.
As a result, agents require more interactions over many time steps to fully switch silos in the GMM model. 

Additionally, in the LLM setting, the silo of each agent is determined by their output at each time step. While this is a necessary estimate of the agent's perspective due to the black-box nature of LLMs, it can be highly variable, especially if the agent holds a multi-modal perspective.  
This is further compounded by the fact that the RAG database may have many sentences with conflicting perspectives. The LLMs internal knowledge may have a preference for a certain flower, while the database may contain information that contradicts this internal preference as well as other information within the RAG. This is an example of imperfect retrieval, which has been shown to hamper the ability of LLMs in knowledge tasks \cite{wang2024astute}. Moreover, this may cause additional unpredictability in LLM output, leading to the appearance of agents that switch more readily between different silos. 

\section{Conclusion}
\label{sec:conclusion}

In this paper, we propose systems of interacting GMMs as a model that can effectively replicate much of the complexity arising from system of interacting LLMs at a fraction of the computational cost,  with the added benefit of mathematical tractability. Comparisons with systems of interacting LLMs demonstrate similarities between the two approaches in the dynamics of mirroring on social alignment and the formation of silos, in particular the existence of unstable silo, and the dependence of siloing effects on model parameters. 
Not only do GMMs as a proxy for LLMs hold promise, their flexibility opens other avenues of research;
our approach can be adjusted to use multivariate Gaussians where any of the parameters are allowed to vary. 
This allows for fine grained evolution of agents in the system, which can better mimic the behaviors of the motivating LLM counterparts. 
Moreover, in our model, we use a simple method to determine silos by looking at the largest weight in the weight vector. However, this may not capture the complexities where agents may have a multi-modal distribution over the possible perspectives. More sophisticated measurement methods can be used such as the use of clustering algorithms on the entire weight vector to provide more accurate silos of the agents. If the means and variances are allowed to vary, this could also be taken into account when determining the similarities between agents. 
Also, the updating mechanism can be made more elaborate: for instance, we can change the RAG set update to follow more conventional models of human memory, taking into account similarity of information to data already in memory as well as the time the information was acquired. When the GMM is updated, we can also weights the information in the RAG set by importance. Such adjustments can also be helpful in reproducing behavior of interacting LLMs for more complex queries and for richer comparisons to an array of human interactions.

\begin{figure}[t!]
    \centering
    \includegraphics[width=1\linewidth]{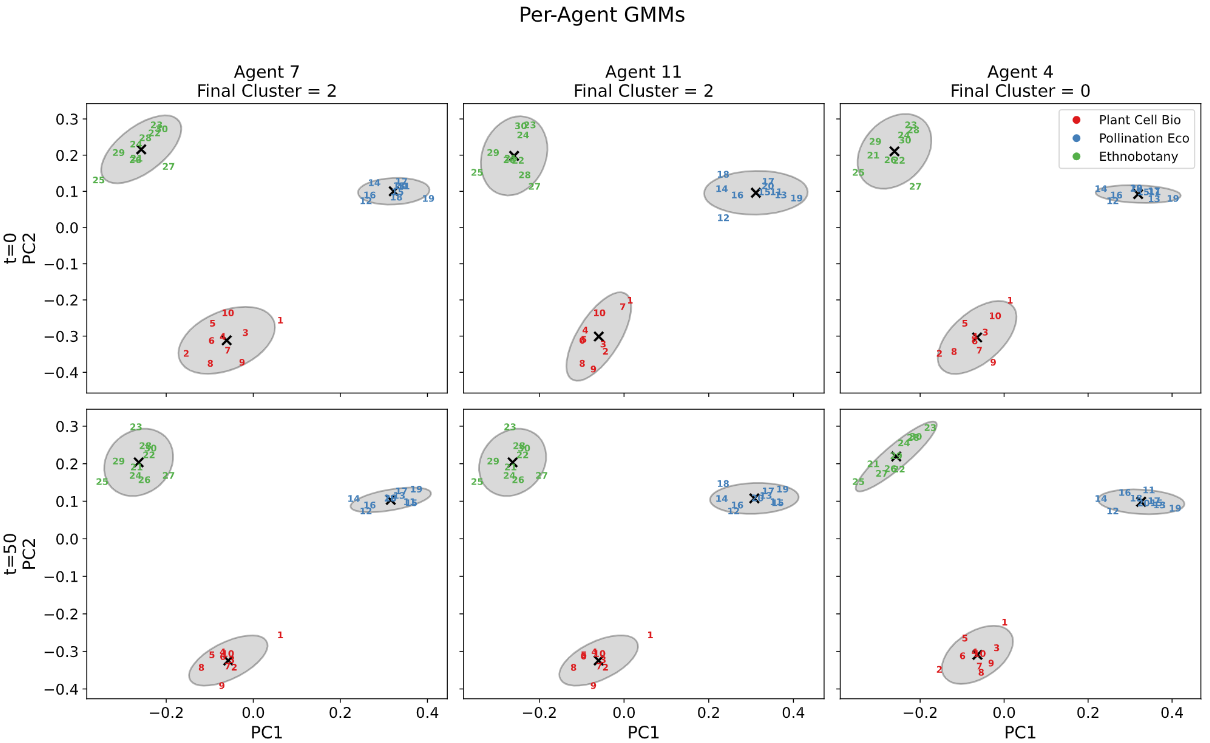}
    \caption{Example sketches of three LLMs in response space, where PCA is used to get a 2-d representation of the LLM sketch.
    The top row shows the sketches at time $t=0$, the bottom row at time $t=50$.
    The first columns are LLMs that ultimately end up in cluster 2 (see Figure \ref{fig:perspective}); the third column an LLM that ends up in cluster 0.
    Here 30 questions---on 3 major topics---were asked of the LLM.}
    \label{fig:llmgmm}
\end{figure}


Insomuch as we can think of the LLM agents as being represented by high-dimensional mixture models (Gaussian or otherwise), our theory on GMM siloing can be extended to prove a similar siloing occurs in LLM space---what was essential to the theory is the idea that agents can move eventually between clusters, though the cluster centers are sufficiently separated to make this movement relatively difficult.
That said, working directly with the LLM models or weighted lattices is daunting.
However, there is evidence that LLM sketches in response space \citep{helm2024statistical} are well-approximated via mixture models; see Figure \ref{fig:llmgmm} for example.
To sketch LLMs in response space, a large number of questions (the same across all models) are asked of a collection of LLMs, with each LLM's answers---appropriately embedded into Euclidean space, for example, via Nomic Embed \cite{nussbaum2024nomic}---effectively producing a sketch of the LLM in Euclidean space. 

\begin{figure}[t!]
    \centering
    \includegraphics[width=1\linewidth]{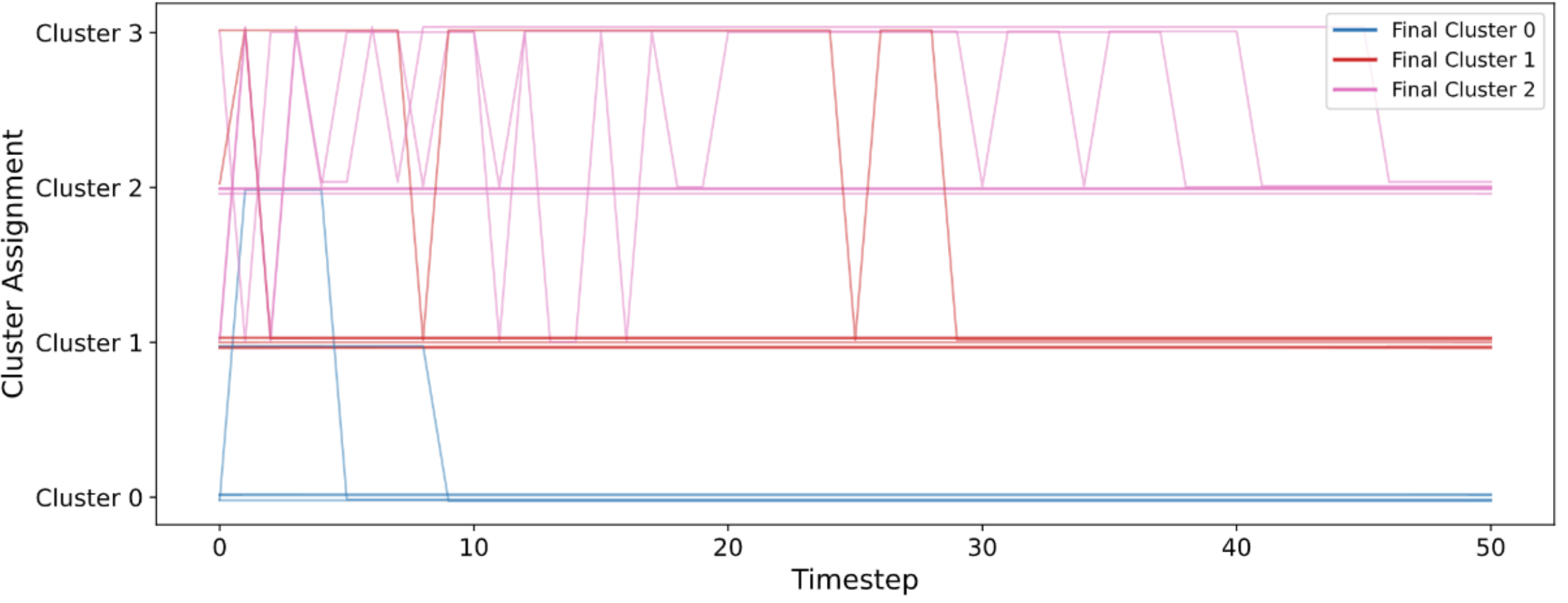}
    \caption{Clustering the collection of LLMs in sketch response space yields stable polarization mirroring the polarization of the interacting LLM and GMM processes.
    Note that this clustering is performed in Data Kernel Perspective Space \citep{helm2024statistical} with Frobenius norm distances between the sketch responses as the chosen distance.
    Here there are:
    $m=20$ agents; 
    $30$ questions (on 3 topics) are used for sketch response space;
    $10$ questions asked per agent interaction (to speed up the interaction process); 
    and
    $p=0.3$ mirroring probability.}
    \label{fig:perspective}
\end{figure}

Moreover, there is evidence that the polarization in the interacting LLM system is manifest in sketch response space as well.
To wit, Figure  \ref{fig:llmgmm} shows the sketch responses of three LLMs in the interacting system at time $t=0$ and after polarization at time $t=50$; two of these LLMs polarize into the same silo, with the third LLM ending in a different silo.
From the figure, we see that these sketches are well-approximated by mixture models with well-separated covariances, and that the siloing appears to manifest as a parameter difference in the fitted GMMs (here in covariance structure; note that the means seem to be fixed in time and across models). 
\emph{If} we can show this rigorously, and \emph{if} our GMM interaction mechanism well-approximates the LLM interaction mechanism in sketch response space, then our theory gives a path forward for proving LLM polarization in sketch response space.
Importantly, recent work has shown that LLM behavior in model space is well-approximated by LLM behavior in the correct sketch response space \cite{acharyya2025concentration}.
These ideas hinge on a (potentially increasingly difficult to prove) sequence 
of ``ifs,'' and while much work is yet to be done to establish these formally, a path forward is at least visible to complete the circle:  we approximate LLMs via GMMs and prove theory for GMMs to elucidate theory for the LLMs as well.

\vspace{2mm}

\noindent\textbf{Acknowledgment:} The authors would like to acknowledge support from the following funding mechanisms: Air Force Office of Scientific Research (AFOSR) Complex Networks award number FA9550-25-1-0128; and the The Johns Hopkins University HLT COE.

\vskip 0.2in

\bibliography{refs}

\newpage
\appendix
\section{Proof of Theorem \ref{thm:polar}}
\label{app:proof1}

Recall the form each of the $GMM$s in the interacting system is $w \mathcal{N}(-1,\sigma^2) + (1-w) \mathcal{N}(1,\sigma^2)$.
Define the function $h_\sigma:[0,1]\times\mathbb{R}\mapsto [0,1]$ via
$$h_\sigma(w,x)=\frac{w}{w+(1-w)e^{2x/\sigma^2}}=\frac{1}{1+\left(\frac{1-w}{w}\right)e^{2x/\sigma^2}};$$
Note that 
\begin{align*}
\frac{\partial}{\partial x}&h_\sigma(w,x)=\frac{-2w(1-w)e^{2x/\sigma^2}}{\sigma^2(w+(1-w)e^{2x/\sigma^2})^2}<0,\quad
\frac{\partial}{\partial w}h_\sigma(w,x)=\frac{e^{2x/\sigma^2}}{(w+(1-w)e^{2x/\sigma^2})^2}>0.
\end{align*}
Let $\rho<1/2$, and recall/define
\begin{align*}
    \eta&=p_{\rho,\sigma}=\mathbb{P}(\mathcal{N}(-1,\sigma^2)\in -1\pm \rho)\\
    C&=C_{\rho,\sigma}=e^{2(1-\rho)/\sigma^2}>e^{1/\sigma^2}\\
    B&=B_{\rho,\sigma}=e^{2(1+\rho)/\sigma^2}<e^{3/\sigma^2}.
\end{align*}
If the current state of the GMM MC at time $t>0$ is $\{(w_i^{(t)},x_{1i}^{(t)},\cdots,x_{ri}^{(t)})\}_{i=1}^m$, the update mechanism for GMM $i$ when incorporating new data $y$ and removing RAG element $x_{ji}^{(t)}$ is given by
\begin{align}
\label{eq:wupdate}
w_i^{(t+1)}=\frac{1}{r}\left(\sum_{\ell=1, \ell\neq j}^r h(w_i^{(t)},x_{\ell i}^{(t)})+h(w_i^{(t)},y)\right).
\end{align}
\begin{prop}
\label{prop:1step}
    Consider a GMM with weight $w\in \mathcal{I}_1$ and RAG $\{x_i\}_{i=1}^r$ where each $x_i\in(1-\rho,1+\rho)\cup(-1-\rho,-1+\rho)$.
    After one received update in $(1-\rho,1+\rho)\cup(-1-\rho,-1+\rho)$, the 
    updated weight is in $\left(\frac{1}{1+CB}, \frac{1}{1+(CB)^{-1}}\right)$.
\end{prop}
\begin{proof}
    Let $w'$ be the updated weight and let $\{x_i'\}_{i=1}^r$ be the updated RAG.  We have that for each $i$, 
    \begin{align*}
       \frac{1}{1+CB}=h\left(\frac{1}{1+C}, 1+\rho\right) \leq h(w,x_i') \leq h\left(\frac{1}{1+C^{-1}}, -1-\rho\right)=\frac{1}{1+(CB)^{-1}}.
    \end{align*}
    Plugging this into Eq. \ref{eq:wupdate} yields the desired result.
\end{proof}

\begin{prop}
\label{prop:manystep}
Consider a GMM with weight $w\in \mathcal{I}_1$ and RAG $\{x_i\}_{i=1}^r$ where each $x_i\in(1-\rho,1+\rho)\cup(-1-\rho,-1+\rho)$.
After $4r+2$ updates, each in $1\pm \rho$ (resp., $-1\pm \rho$) the updated weight is at most $\frac{1}{1+C}$ (resp., at least $\frac{1}{1+C^{-1}}$).
\end{prop}
\begin{proof}
    Applying Proposition \ref{prop:1step} for $r$ steps yields that after $r$ updates, the weight of the GMM is in $\left(\frac{1}{1+CB^k}, \frac{1}{1+C^{-1}B^{-k}}\right)$.
    We will deal with the case where all updates came from $1\pm \rho$, the case where they came from $-1\pm \rho$ being analogous.
    If all these updates came from $1\pm \rho$, then after an additional $3r+2$ updates in $1\pm \rho$, the weight is \emph{at most}
    \begin{align*}
    h&\left(\cdots h\left(h\left(h\left( \frac{1}{1+C^{-1}B^{-r}}, 1-\rho\right),1-\rho\right),1-\rho\right)\cdots,1-\rho\right)\\
    &=\frac{1}{1+C^{3r+2-1}B^{-r}}=\frac{1}{1+C\left(\frac{C^3}{B}\right)^r}<\frac{1}{1+C}
    \end{align*}
    as desired.
\end{proof}
\noindent Note that as the update is monotonically increasing in $w$, an immediate extension of Proposition \ref{prop:manystep} is that if the GMM has weight $w < \mathcal{I}_1$ (resp., $w > \mathcal{I}_1$) and RAG $\{x_i\}_{i=1}^r$ where each $x_i\in(1-\rho,1+\rho)\cup(-1-\rho,-1+\rho)$, then after 
$4r+2$ updates, each in $1\pm \rho$ (resp., $-1\pm \rho$) the updated weight again is at most $\frac{1}{1+C}$ (resp., at least $\frac{1}{1+C^{-1}}$).

Our first main result, Theorem \ref{thm:polar} gives a uniform lower bound on the probability of the system polarizing after a fixed number of steps.
Before proving Theorem \ref{thm:polar}, we first need to establish conditions that ensure the RAG behaves well enough to then polarize.

\subsection{Resetting the RAG}
\label{sec:reset}

Our lower bound on polarization depends on there not existing extreme elements in the RAGs of our interacting GMMs.
This is captured by the notion of a \emph{well-behaved RAG} as defined below.
\begin{defn}
\label{def:behave}
     We say that the interacting GMM system at time $t$, denoted via 
     $$\mathfrak{G}_t=\{(w_i^{(t)},x_{1i}^{(t)},\cdots,x_{ri}^{(t)})\}_{i=1}^m,$$ has a \emph{well-behaved RAG} if $\left\{\{x_{\ell i}^{(t)}\}_{\ell=1}^r\right\}_{i=1}^m\subset (1\pm \rho)\cup (-1\pm \rho)$; i.e., all RAG elements at time $t$ are in $(1\pm \rho)\cup (-1\pm \rho)$.  
     At a time $t$, let the event $\mathcal{B}^{(t)}$ be the event that the interacting GMM system is well-behaved at time $t$.
\end{defn}
\noindent Note that if the system is level-$\ell$ polarized for $\ell\geq 1$, then the RAG is well-behaved by definition.

The following Lemma gives a lower bound for the interacting GMM system to have a well-behaved RAG at time $t+mr$ independent of the state of the system at time $t$.
To wit we have the following.
\begin{lem}
\label{lem:behave}
    Let $\xi=\xi_{\rho,r,\sigma}=\mathbb{P}(\mathcal{N}(1,\sigma^2)\in 1\pm \frac{\rho}{2r})$.  
    We have that 
    \begin{align*}
    \mathbb{P}&\left(\mathcal{B}^{(t+rm)}\,\big|\, \text{state of GMM system at time t is }\{(w_i^{(t)},x_{1i}^{(t)},\cdots,x_{ri}^{(t)})\}_{i=1}^m
    \right)\\
    &\geq  \frac{m!}{m^{mr}} \left(\frac{m/2-1}{m} \xi\right)^{rm}.
    \end{align*}
\end{lem}
\noindent The proof of Lemma \ref{lem:behave} will make use of the following Proposition, which gives sufficient conditions for a single GMM to have a well-behaved RAG after $k$ update steps. 
\begin{prop}
\label{prop:behave}
Consider a GMM with weight $w$ and RAG $\{x_i\}_{i=1}^r$.  After this GMM receives exactly $r$ updates in $1\pm \frac{\rho}{2r}$ (resp., $-1\pm \frac{\rho}{2r}$), all RAG elements are in $1\pm \rho$ (resp., $-1\pm \rho$).
\end{prop}
\begin{proof}
To ease notation, let $s=\frac{\rho}{2r}$.
We will consider the $1\pm s$ case, with the $-1\pm s$ case following analogously.
    We will show that for all $\ell\leq r$, after exactly $\ell$ updates all in $1\pm s$, there are at least $\ell$ points in $1\pm 2\ell s$.  To see this, we induct on $\ell$.  
    The base case where $\ell=1$ is trivial.  
    Assuming the result holds for an $\ell<r$, consider the $(\ell+1)$st update in $1\pm s$.  
    By the inductive hypothesis, when the $(\ell+1)$st update arrives---denote this point via $y$---there are at least $\ell$ points in $1\pm 2\ell s$; denote these points via $C_\ell$.
    We consider two cases:
    \begin{itemize}
        \item[i.] The point removed from the RAG at time $\ell+1$ is not in $C_\ell$; then there are at least $|C_\ell|+1$ points in $1\pm 2\ell s$ and the result follows immediately. 
        \item[ii.] The point removed from the RAG at time $\ell+1$ is in $C_\ell$.
        As all points in $C_\ell$ are at most distance $2\ell s+s$ from the newly arriving update $y$ and only the farthest point from $y$ is removed, all non-removed RAG elements can be distance at most $2\ell s+s$ from $y$.
        Hence, all non-removed RAG elements would need to be in $1\pm(2\ell s+2s)=1\pm 2s(\ell+1)$ as desired.
    \end{itemize}
\end{proof}
\noindent We are now ready for the proof of Lemma \ref{lem:behave}.
\begin{proof}[Proof of Lemma \ref{lem:behave}]
Consider the state of the interacting GMM system at time $t$ being given by $\{(w_i^{(t)},x_{1i}^{(t)},\cdots,x_{ri}^{(t)})\}_{i=1}^m$.
We will provide a sequence of events that is sufficient to ensure a globally well-behaved RAG $mr$ steps into the future. To wit, if we observe
\begin{itemize}
\item[i.] There is a permutation of $[m]$, denoted $\tau$, such that the the interacting models chosen to be updated are ordered via
$$\underbrace{\tau(1),\cdots,\tau(1)}_{\text{updated r times}},
\underbrace{\tau(2),\cdots,\tau(2)}_{\text{updated r times}},
\cdots,
\underbrace{\tau(m),\cdots,\tau(m)}_{\text{updated r times}};$$
the probability of this occurring is at least 
$
\frac{m!}{m^{mr}}.
$
\item[ii.] After the $rh$-th update for $h=0,\ldots,m-1$, let the total weight put on the $\mathcal{N}(1,\sigma^2)$ Gaussian be denoted $W^{(rh)}_+$ (with  $W^{(rh)}_-$ the weight on the $\mathcal{N}(-1,\sigma^2)$ Gaussian).  Since $W^{(rh)}_++W^{(rh)}_-=m$, we have that at least one of $W^{(rh)}_+$, $W^{(rh)}_-$ is at least $m/2$.
We consider two cases.
\begin{itemize}
    \item[a.] If $W^{(rh)}_+\geq m/2$, we then have that all updates to the $\tau(h+1)$st model are in $1\pm \frac{\rho}{2r}$; this ensures by Proposition \ref{prop:behave} that all RAG elements for this model are in $1\pm \rho$ after the $r$ updates.
    In this case, the total weight on $\mathcal{N}(1,\sigma^2)$ for each of these updates is at least $(m/2-1)$ (as throughout these updates, only $w_{\tau(h+1)}$ is being updated and this can impact the total weight on the $\mathcal{N}(1,\sigma^2)$ Gaussian by at most 1).
    \item[b.] If $W^{(kh)}_-\geq m/2$ and $W^{(kh)}_+< m/2$, we then have that all updates to the $\tau(h+1)$st model are in $-1\pm \frac{\rho}{2r}$; this ensures by Proposition \ref{prop:behave} that all RAG elements for this model are in $-1\pm \rho$ after the $r$ updates. 
    In this case, the total weight on $\mathcal{N}(-1,\sigma^2)$ for each of these updates is at least $(m/2-1)$.
\end{itemize}
Therefore, a lower bound for this occurring for all $h$ is at least $\left(\frac{m/2-1}{m} \xi\right)^{rm}.$
\end{itemize}
Combining the above two probabilities, we arrive at our desired result.
\end{proof}

\subsection{Polarizing}
\label{sec:polar}
We next consider the probability of polarizing after a fixed number of steps.  Note that our strategy is to first ensure the updates provide a well-behaved RAG and then consider polarization.
We suspect that the steps ensuring the RAG is well-behaved are extraneous, though dealing with arbitrary RAG elements gets cumbersome for general RAG sizes.

\begin{lem}
\label{lem:pol}
We have that 
    \begin{align*}
            \mathbb{P}&\left(E_1^{(t+(4r+2)m)}\,\big|\,\text{state of GMM MC at time }t\text { is } \{(w_i^{(t)},x_{1i}^{(t)},\cdots,x_{ri}^{(t)})\}_{i=1}^m\in \mathcal{B}^{(t)}\right)\\
            &\quad\quad\quad\geq \frac{((4r+2)m)!}{((4r+2)!)^m m^{(4r+2)m}}\left(\frac{\lfloor m/2\rfloor}{m} \frac{\eta}{1+CB^r}\right)^{(4r+2)m}
    \end{align*}
    holds for any $\{(w_i^{(t)},x_{1i}^{(t)},\cdots,x_{ki}^{(t)})\}_{i=1}^m\in \mathcal{B}^{(t)}$ in the GMM MC state space.
\end{lem}
\begin{proof}
For the configuration 
$\{(w_i^{(t)},x_{1i}^{(t)},\cdots,x_{ri}^{(t)})\}_{i=1}^m\in\mathcal{B}^{(t)}$, 
define the sets 
$$
L_t=\{i:w_i^{(t)}<\mathcal{I}_1\};\quad 
C_t=\{i:w_i^{(t)}\in\mathcal{I}_1\};\quad 
R_t=\{i:w_i^{(t)}>\mathcal{I}_1\}.
$$
Now, observe that each index in $L_t\cup C_t$ or (resp., $R_t\cup C_t$) will be in $L_t\cup C_t$ (resp., $R_t\cup C_t$) after $4r+2$ updates all in $1\pm \rho$ (resp., $-1\pm \rho$) by Proposition \ref{prop:manystep}. 
Therefore if 
\begin{itemize}
    \item[i.] All GMM's indexed in $L_t\cup C_t$ receive exactly $4r+2$ updates in $1\pm \rho$, and each of these updates comes from another GMM in $L_t\cup C_t$;
    \item[ii.] All GMM's indexed in $R_t$ receive exactly $4r+2$ updates in $-1\pm \rho$, and each of these updates comes from another GMM in $R_t$;
\end{itemize}
then the GMM system is \emph{level-1} polarized (this would be after $(4r+2)m$ steps).
The probability of the above occurring is lower bounded by 
\begin{align*}
 &\underbrace{\frac{((4r+2)m)!}{((4r+2)!)^m m^{(4r+2)m}}  }_{\substack{\text{each GMM updated}\\\text{exactly 4r+2 times}}}\bigg(\underbrace{\left(\frac{|C_t\cup L_t|}{m}\right)^{4r+2}}_{\substack{\text{Each $C_t\cup L_t$ update}\\\text{chosen from }C_t\cup L_t}} \underbrace{\left[\eta\left(1-\frac{1}{1+C^{-1}B^{-r}}\right)\right]^{4r+2}}_{\substack{\text{Each $C_t\cup L_t$ update}\\\text{is in }1\pm \rho}} \bigg)^{|C_t\cup L_t|}\times\cdots \\
&\quad \cdots\times \bigg(\underbrace{\left(\frac{|R_t|}{m}\right)^{4r+2}}_{\substack{\text{Each $R_t$ update}\\\text{chosen from }R_t}} \underbrace{\left(\frac{\eta}{1+CB^r}\right)^{4r+2}}_{\substack{\text{Each $R_t$ update}\\\text{is in }1\pm \rho}} \bigg)^{|R_t|}\\
&\geq  \frac{((4r+2)m)!}{((4r+2)!)^m m^{(4r+2)m}}\left(\frac{\lfloor m/2\rfloor}{m} \frac{\eta}{1+CB^r}\right)^{(4r+2)m}
\end{align*}
where the last inequality followed from 
$|L_t\cup C_t|=m-|R_t|$ and the function
$$
f(x)=\left(x\right)^x\left(m-x\right)^{m-x}\geq (\lfloor m/2\rfloor)^m
$$
for $x$ in the range $[1,m]\cap\mathbb{Z}$.
Noting this lower bound is independent of the initial configuration yields the desired result.
\end{proof}
Combining Lemmas \ref{lem:behave} and \ref{lem:pol}, we arrive at Theorem \ref{thm:polar}, restated here for clarity.

\vspace{2mm}
\noindent \textbf{Theorem \ref{thm:polar}}
\emph{There is a constant $q=q_{m,\rho,r}>0$ such that 
    $$
    \mathbb{P}\left(E_1^{(t+(5r+2)m)}\,\big|\,\text{state of GMM MC at time }t\text { is } \{(w_i^{(t)},x_{1i}^{(t)},\cdots,x_{ri}^{(t)})\}_{i=1}^m\right)\geq q
    $$
    holds for any $\{(w_i^{(t)},x_{1i}^{(t)},\cdots,x_{ri}^{(t)})\}_{i=1}^m$ in the GMM MC state space.}\\
\begin{proof}
    Let the GMM system at time $t$ be denoted $\mathfrak{G}^{(t)}$. Then
    \begin{align*}
        \mathbb{P}&\left(E_1^{(t+(5r+2)m)}\,\big|\mathfrak{G}^{(t)}=\{(w_i^{(t)},x_{1i}^{(t)},\cdots,x_{ri}^{(t)})\}_{i=1}^m\right)\\
        &\geq\mathbb{P}\left(E_1^{(t+(5r+2)m)}\,\big|\mathfrak{G}^{(t)}=\{(w_i^{(t)},x_{1i}^{(t)},\cdots,x_{ri}^{(t)})\}_{i=1}^m, \mathcal{B}^{(t+mr)}\right)\times\cdots\\
        &\quad \cdots\times\mathbb{P}\left( \mathcal{B}^{(t+mr)} \big| \mathfrak{G}^{(t)}=\{(w_i^{(t)},x_{1i}^{(t)},\cdots,x_{ri}^{(t)})\}_{i=1}^m\right)\\
        & \geq  \frac{((4r+2)m)!}{((4r+2)!)^m m^{(4r+2)m}}\left(\frac{\lfloor m/2\rfloor}{m} \frac{\eta}{1+CB^r}\right)^{(4r+2)m} \frac{m!}{m^{mr}} \left(\frac{m/2-1}{m} \xi\right)^{rm}
    \end{align*}
    where the last inequality follows from the Markov property.
    Since this bound is independent of the initial state, the result follows.
\end{proof}

\section{Results of the Experiment with Variable Standard Deviation}
\label{app:exp_std}
This experiment is in the same setting and initialization as the experiment in Section \ref{sec:results}. The only difference is that now the covariance matrix \(\{\mathbf{V}_{i}^{(t)}\}\) is also a variable changing over time. The Algorithm \ref{alg:standard-simulation-procedure} would change slightly to be compatible with changing covariance. Now, the initial covariance matrices \(\{\mb{V}_i^{(0)}\}_{i=1}^m\) are given as input to Algorithm \ref{alg:variance-simulation-procedure} and in each iteration we also update them for each agent based on the M-step in the EM algorithm. The helper function \texttt{UpdateGMMCov} takes the same inputs as \texttt{UpdateGMM}, but returns a tuple consisting of the updated weights and the updated covariances. 

\begin{algorithm}[t!]
\caption{Simulation procedure with variable covariance matrix} 
\label{alg:variance-simulation-procedure}
\begin{algorithmic}[1]
    \Procedure{SimulateInteractingGMMsWithVariableCovariance}{$T, p, k, r, n, m,\{\mb{w}_i^{(0)}\}_{i=1}^m, \mb{M}, \{\mb{V}_i^{(0)}\}_{i=1}^m$, $\epsilon>0$}
    \For{$i \in \{1,\dots, m\}$}\Comment{Initialization}
    \State $R_i^{(0)} \gets $\texttt{SampleFromGMM}($\mb{w}_i^{(0)},\mb{M}, \mb{V}_i^{(0)}$, $r$)
    \EndFor
    \For{$t=1,2, \dots, T$} \Comment{Interaction}
    \For{$i \in \{1, \dots, m\}$}
        \State $u \gets $\texttt{Uniform}(0,1)
        \If{$u < p$}
            \State $j \gets i$
        \Else
            \State $\mathcal{N}_i\gets $\texttt{GetKNearestGMMs}$\big(k, i ,t, \{\mb{w}_{\ell}^{(t)}\}_{\ell=1}^m\big)$ 
            \State $j$ is uniformly chosen from $\mathcal{N}_i$
        \EndIf
        \State $x \gets$ \texttt{SampleFromGMM}($\mb{w}_i^{(t-1)},\mb{M},\mb{V}_i^{(t-1)}$, 1)\Comment{Query}
\State $\tilde R_j^{(t)}\gets (R_j^{(t)} \cup \{x\}) \setminus (\arg\max_{v \in R_j^{(t)}} \|x-v\|)$\Comment{Pseudo-update}
\State $\tilde {\mb{w}}_j^{(t)}, \tilde {\mb{V}}_j^{(t)}\gets$ \texttt{UpdateGMMCov}$(\tilde R_j^{(t)},\mb{M},\mb{V}_j^{(t-1)},\mb{w}_j^{(t-1)}$)
\State y $\gets$ \texttt{SampleFromGMM}($\tilde {\mb{w}}_j^{(t)},\mb{M},\tilde {\mb{V}}_j^{(t)},1$)\Comment{Answer}
\State $R_{i}^{(t+1)} \gets (R_i^{(t)} \cup \{y\}) \setminus (\arg\max_{v \in R_i^{(t)}} \|y-v\|)$\Comment{RAG update}
\State $\mb{w}_i^{(t+1)} , \mb{V}_i^{(t+1)} \gets$ \texttt{UpdateGMMCov}$(R_i^{(t+1)},\mb{M},\mb{V}_i^{(t)},\mb{w}_i^{(t)})$\Comment{Model update}
        \EndFor
    \EndFor
    \EndProcedure
\end{algorithmic}
\end{algorithm}

\begin{figure}[t!]
    \centering
    \includegraphics[width=1\textwidth]{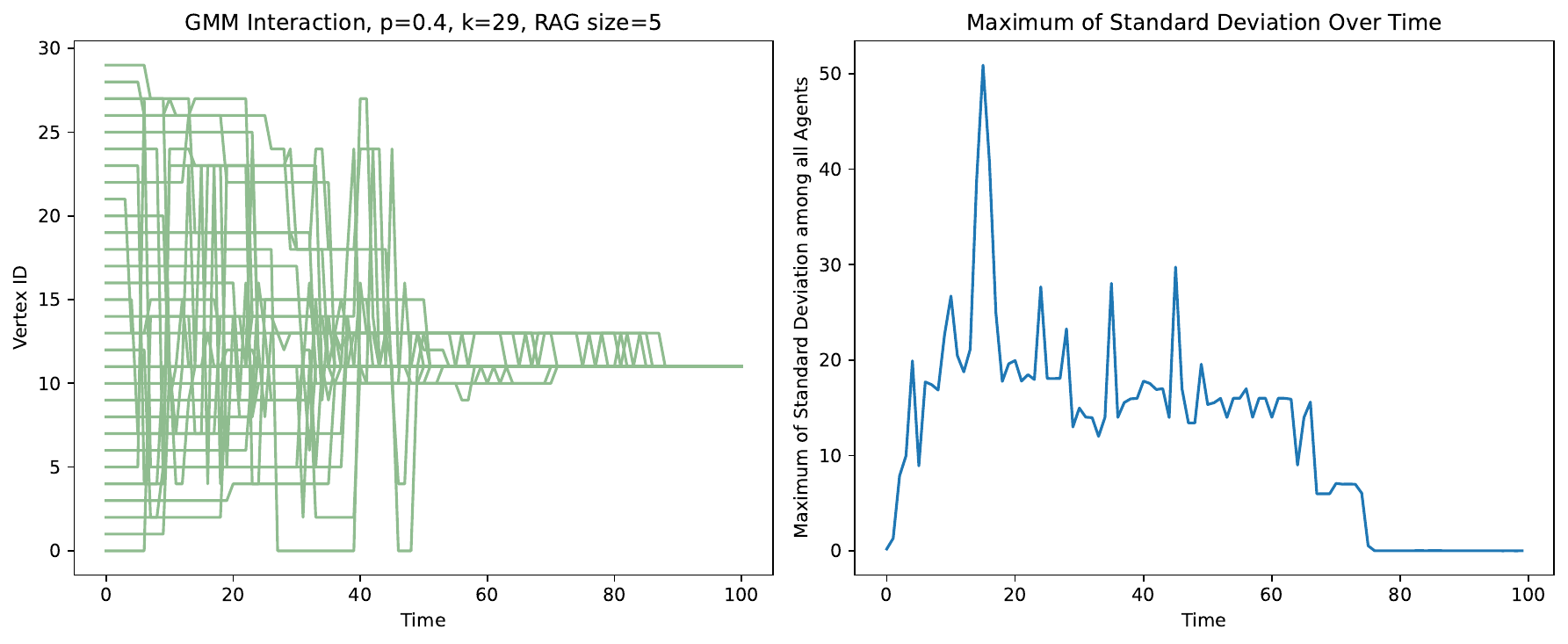}
    \caption{\textbf{(Left)} Vertex ID assignments for each agent, demonstrating the convergence of all agents to a single Gaussian component (silo) as time progresses. \textbf{(Right)} The maximum standard deviation among all agents over time. Initial spikes in standard deviation occur when distant RAG elements are assigned to a component; however, as the RAG elements are reassigned to closer components or removed, the maximum standard deviation eventually decreases to zero.}
    \label{fig:unstable-silo-var}
\end{figure}

As you can see in the left plot of Figure \ref{fig:unstable-silo-var}, we observe the same behavior as in Section \ref{sec:results} in terms of siloing behavior.
Notably, the right plot of Figure \ref{fig:unstable-silo-var} shows that that the maximum of all the 900 standard deviations eventually decreases to zero. Some of the standard deviations update to relatively large values because a far RAG element is assigned to them, causing their standard deviation to grow. However, after a few iterations, those RAG elements either get removed from the RAG, or assigned to a closer Gaussian component. 

\end{document}